\documentclass[10pt,twocolumn,letterpaper]{article}
\usepackage[accsupp]{axessibility} 
\usepackage{cvpr}              
\usepackage{xspace}
\usepackage{amsmath}
\usepackage{algorithm}
\usepackage{algorithmic}
\usepackage{multirow}
\usepackage{makecell}
\def\onedot{.\xspace}

\newcommand{\hut}{\textcolor{black}}
\newcommand{\yr}{\textcolor{black}} 
\newcommand{\crr}{\textcolor{black}} 
\newcommand{\yl}[1]{{\textcolor{black}{#1}}} 

\usepackage[dvipsnames]{xcolor}

\definecolor{cvprblue}{rgb}{0.21,0.49,0.74}
\usepackage[pagebackref,breaklinks,colorlinks,citecolor=cvprblue]{hyperref}

\def\paperID{1637} 
\def\confName{CVPR}
\def\confYear{2024}

\title{SuperSVG: Superpixel-based Scalable Vector Graphics Synthesis}

\author{Teng Hu$^{1}$,~~~Ran Yi$^{1}$\thanks{Corresponding author.},~~~Baihong Qian$^{1}$,~~~Jiangning Zhang$^{2}$,~~~Paul L. Rosin$^{3}$,~~~ Yu-Kun Lai$^{3}$\\
$^1$Shanghai Jiao Tong University,~~~$^2$Youtu Lab, Tencent,~~~$^3$Cardiff University\\
$\{$hu-teng, ranyi, cherry\_qbh$\}$@sjtu.edu.cn,~~~
vtzhang@tencent.com,~~~
$\{$RosinPL, LaiY4$\}$@cardiff.ac.uk
}

\begin{document}
\maketitle
 \begin{abstract}

SVG (Scalable Vector Graphics) \yr{is} a widely used graphics format \yr{that} possesses excellent scalability and \yr{editability}. 
\yr{Image vectorization, which aims \yl{to convert} raster images to SVGs, is an important yet challenging problem in computer vision and graphics.}
Existing \yr{image vectorization} methods either suffer from low reconstruction accuracy \yr{for complex images} 
or require \yr{long} computation time. 
To address this issue, we propose SuperSVG, a superpixel-based vectorization model that achieves fast and high-precision image vectorization. 
Specifically, we decompose the input image into superpixels to help the model focus on areas with similar colors and textures. 
Then, we propose a two-stage self-training framework, where a coarse-stage model is employed to reconstruct the \yr{main} structure and a refinement-stage model is used for enriching the details. 
Moreover, we propose a novel dynamic path warping loss to help the refinement-stage model to inherit knowledge from the coarse-stage model. 
Extensive qualitative and quantitative experiments demonstrate the superior performance of our method in terms of reconstruction accuracy and inference time compared to \yr{state-of-the-art} approaches.
The code is available in
\url{https://github.com/sjtuplayer/SuperSVG}.

\end{abstract}    
\section{Introduction}
\label{sec:intro}

Scalable Vector Graphics, commonly known as SVG, is a widely used vector image format that 
\yr{has} a wide range of applications and advantages within the domains of web design, graphic design, mobile applications, data visualization, and various other contexts. Compared with raster \yl{images} that \yl{represent content} by pixels, SVG describes images by parameterized vectors and benefits from its scalability and 
\yr{editability}
where it can be resized to any \yl{resolution} without losing quality and can be easily manipulated by its layer-wise topological information.

\begin{figure}[t]
\centering
\includegraphics[width=0.44\textwidth]{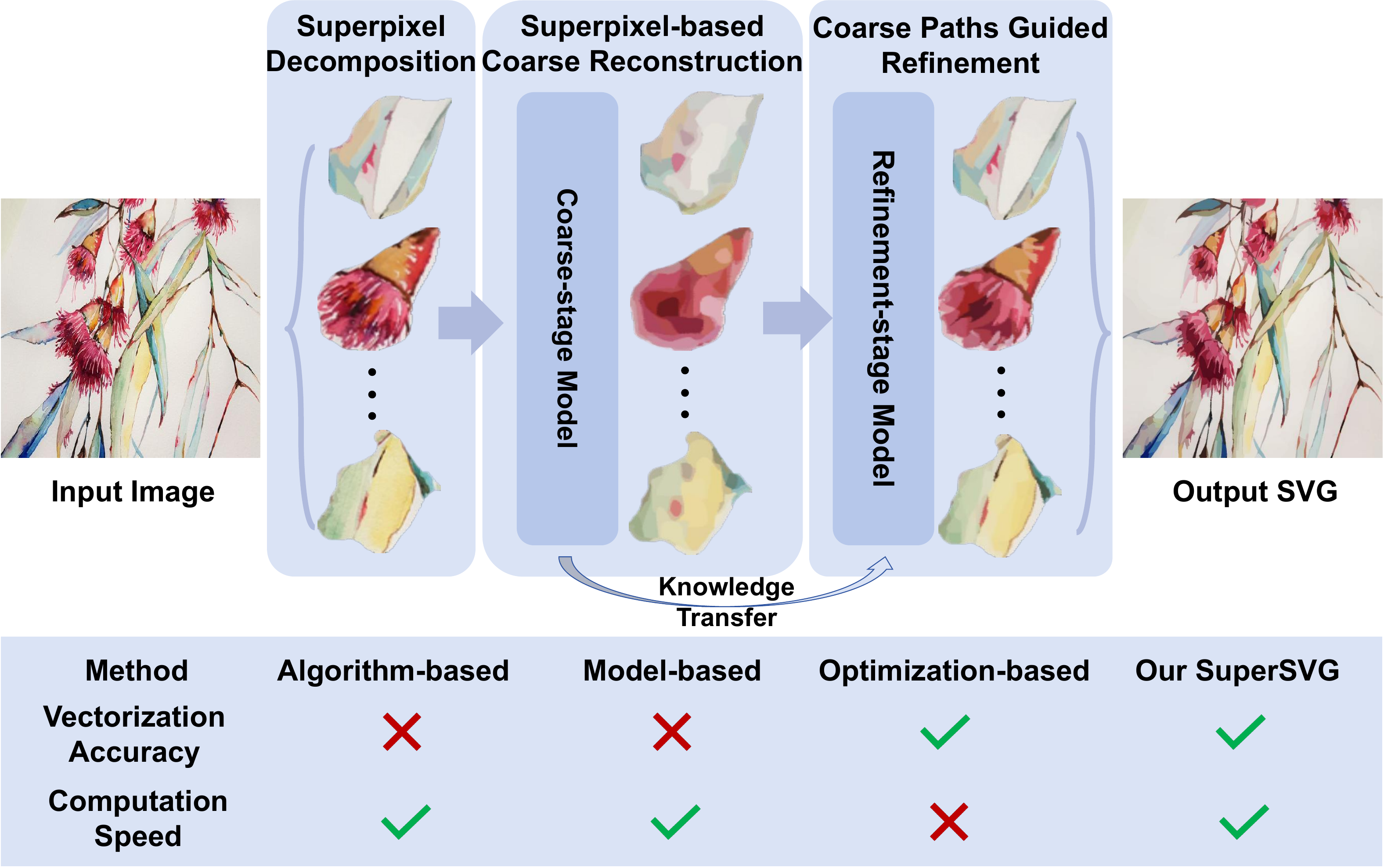}
\vspace{-0.1in}
\caption{Overview of our SuperSVG: our model first decomposes the image \yl{to be vectorized} into superpixels, 
\yl{each containing}
pixels \yl{sharing} similar colors and contents. The \yr{coarse-stage} model predicts the path parameters \yr{to reconstruct the main structure, }
and then the 
\yr{coarse paths guided} \yr{refinement} model enriches the details by learning the knowledge from the 
\yr{coarse-stage}
model. 
Compared to the previous methods, our SuperSVG achieves both a high vectorization accuracy and fast computation speed.}
\label{fig:motivation}
\vspace{-0.15in}
\end{figure}

Given the superior capabilities of Scalable Vector Graphics (SVG) in image representation and editing, there is much research on the topic of image vectorization, which aims to convert rasterized images into SVG. The existing methods can be categorized into three classes: 
\textit{\textbf{1)} Traditional algorithm-based methods}~\cite{potrace,curvilinear, subdivision, topology-preserving, bayesianIV, clipart}, where conventional algorithms are employed to fit images, \hut{but they usually suffer from a lower vectorization quality.} 
\textit{\textbf{2)} Deep-learning-based methods}~\cite{im2vec, raster2vec, deepspline, sketch, sketchformer,hu2023stroke}, which parameterize raster images using deep neural networks for reconstruction. 
\yr{\yl{They} \yl{are efficient and} can handle the vectorization of simple graphics or characters ({\it e.g.,} icons and emojis), but }
\yl{struggle}
to vectorize complex images. 
\textit{\textbf{3)} Optimization-based methods}~\cite{diffvg,live,zhu2023samvg,curvedSBR}, which optimize SVG parameters to fit the target image, yielding relatively superior reconstruction quality. 
However, these methods entail a substantial amount of time and computational resources, making them impractical for timely processing of large-scale data. 
In summary, previous image vectorization methods 
\yr{either suffer from}
low reconstruction quality \yr{for complex images}, 
or demand extensive computation time, imposing significant constraints on their practical utility.

\hut{To achieve good \yl{vectorization} quality with high efficiency, we propose {\it SuperSVG}, a deep-learning-based method that translates \yl{images} into scalable vector graphics (SVG) in a coarse-to-fine manner.
\yr{Since neural networks have difficulties in directly vectorizing}
complex images, 
we decompose the input image into different parts \yl{in the form of superpixels} wherein the pixels share similar colors and textures \yr{and then vectorize each part.}
Then, we propose a {\it Two-Stage \yl{Self-Teaching} 
Training framework} to vectorize the superpixels, where the coarse-stage model is trained to reconstruct the \yr{main} structure of the image and the \yl{refinement}-stage model is trained to enrich the image details. 
We make use of the \yr{predicted paths} from the coarse-stage model to \yr{guide} the \yl{refinement}-stage model in image \yl{vectorization}. 
\yr{Furthermore, w}e propose a novel {\it Dynamic Path Warping loss} which helps the \yl{refinement}-stage to inherit the knowledge of the coarse-stage model. 
With the help of the superpixel-based image decomposition and the two-stage self-teaching framework, our SuperSVG can 
\yr{keep the image structure well}
and reconstruct more details at high speed. \yr{Extensive} quantitative and qualitative experiments validate the effectiveness of our model. 
}

The main contributions of our work are \yr{four}-fold:

\begin{itemize}
    \item We propose SuperSVG, a novel \yr{superpixel-based vectorization} model that translates the rasterized images into scalable vector graphics (SVG) based on superpixels and \yr{vectorizes} the superpixels in a coarse-to-fine manner.


    \item We design a 
    \yr{Two-Stage Self-Teaching} Training framework, where we employ a coarse-stage model to reconstruct the \yr{main} structures and a refinement-stage model to enrich the image details based on the coarse-stage output.

    \item We \yr{propose} a 
    \yr{Coarse Paths Guided}
    Training strategy to guide the \yl{refinement}-stage model to inherit the knowledge from the coarse-stage model, which greatly improves the performance of the \yl{refinement}-stage model and 
    \yr{avoids} converging to \yr{suboptimal} local 
    \yr{minimum}.

    \item We propose Dynamic Path Warping (DPW) loss, which measures the distance between 
    \yr{the predicted paths from the refinement-stage model and the pseudo ground truth approximated with coarse paths.}
    By minimizing the DPW loss, the \yl{refinement}-stage model can distill the knowledge from the coarse-stage model.
    
\end{itemize}

\section{Related Work}




\subsection{Image Vectorization}

\hut{Image vectorization aims \yl{to transform} a rasterized image into scalable vector graphics (SVG) composed of parameterized vectors. 
Different from \yr{raster} images that may 
\yl{become blurry}
\yl{when} zooming in, SVG can be rendered at any \yr{resolution without losing} quality and is convenient to edit, widely used in web design, graphic design, etc. The existing vectorization methods can be classified into 3 categories:}

\textbf{Traditional Algorithm-based \yr{Image V}ectorization Methods} 
\hut{can be classified into mesh-based and curve-based ones. 
The mesh-based methods~\cite{curvilinear, subdivision, topology-preserving,bayesianIV} segment \yl{an} input image into non-overlapping patches, and infer the color and the boundary location for each region. 
The curve-based methods~\cite{inverse-diff, hierarchical-diff, autoIV, potrace,AdobeIllustrator} employ B\'ezier curves with different colors defined on either side to create the \yl{vector} image. Potrace~\cite{potrace} is a representative 
\yl{method of this type}
that projects the smooth outlines into B\'ezier paths, and merge the adjacent paths together. However, the vectorization quality of these methods still needs improvements.}
 

\textbf{Deep-learning-based \yr{Image V}ectorization Methods}
\hut{\yl{use} neural networks to project a raster image into SVG. Im2Vec~\cite{im2vec} employs \yl{a} variational auto-encoder \yl{(VAE)}~\cite{vae} to embed the input image and then \yl{maps} it into path parameters by \yl{a Long Short-Term Memory (LSTM) module}~\cite{schuster1997lstm}. Raster2Vec~\cite{raster2vec} is focused on vectorization of rasterized floor plans using a ResNet~\cite{resnet}. Gao et al.~\cite{deepspline} rely on a pre-trained VGG network~\cite{vgg} and \yl{a} hierarchical \yl{Recurrent Neural Network (RNN)} to output parametric curves of different sizes. But these methods only focus on simple images \yl{and} cannot vectorize complex images well. In \yl{comparison}, our SuperSVG is the first deep-learning-based method that can \yl{vectorize} images with \yl{complex details, thanks to our superpixel decomposition and coarse-path guided refinement that substantially reduce the learning difficulties.}}

\textbf{Optimization-based \yr{Image V}ectorization Methods}. \hut{DiffVG~\cite{diffvg}
proposes a differentiable renderer that renders the SVG parameters into images. Based on \yl{this}, DiffVG minimizes the distance between the rasterized and vector images by optimizing the SVG parameters using gradient descent.
LIVE~\cite{live} and SAMVG~\cite{zhu2023samvg} \yl{further} introduce a layer-wise optimization framework, which achieves better vectorization quality over the previous methods. However, due to the low optimization efficiency, they suffer from a long optimization time. In contrast, our SuperSVG achieves both a good vectorization quality and high efficiency.}

\subsection{Superpixel Decomposition}
\hut{Superpixel decomposition is usually used for data preprocessing in vision tasks. Existing superpixel decomposition methods can be categorized into methods based on 
\yr{traditional algorithm}
or deep learning. For the 
\yr{traditional algorithm based}
methods, diverse strategies have been employed, {\it e.g.,} energy-driven sampling~\cite{van2012seeds}, geometric flows~\cite{turbopixel} and clustering~\cite{slic}. Some recent works~\cite{saal,ssn,ye2019fast} employ neural networks to enhance the performance in superpixel decomposition, which shows great potential in this task.}

\begin{figure*}[t]
\centering
\includegraphics[width=0.9\textwidth]{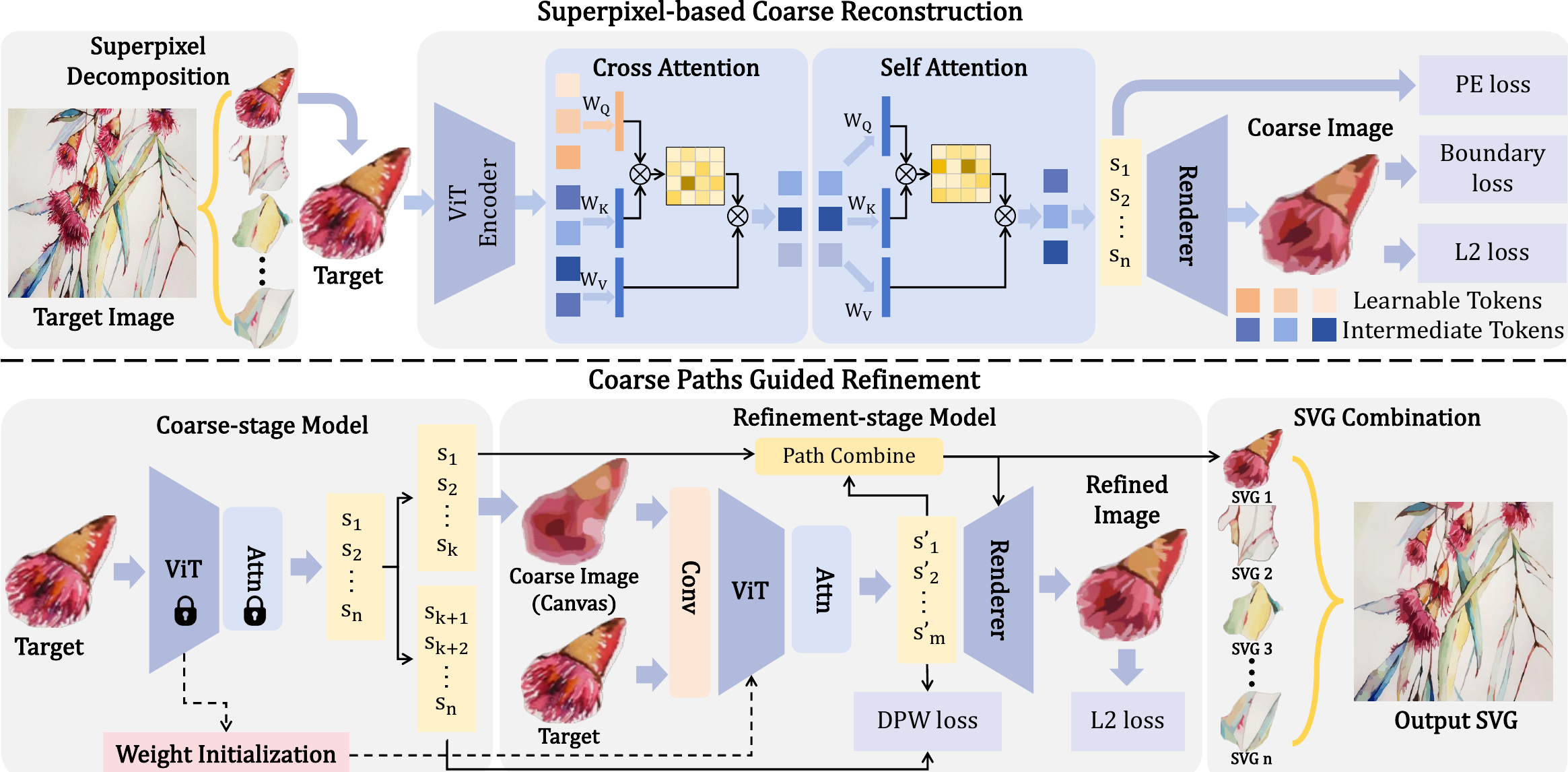}
\vspace{-0.1in}
\caption{\textbf{Main framework of our SuperSVG:} we decompose the target image into superpixels and vectorize each superpixel separately. We employ an attention-based coarse-stage model to \yr{predict SVG paths that} reconstruct the main structure of the superpixel. Then, a refinement-stage model guided by the coarse paths is designed to predict more \yr{SVG paths to refine} details based on the coarse image. Finally, by combining all the predicted SVGs for each superpixel, we \yr{obtain} an output SVG with good structure and fine details. }
\label{fig:main framework}
\vspace{-0.15in}
\end{figure*}

\section{Method}

\hut{Image vectorization aims to translate \yl{a} rasterized image~$I$ into a Scalable Vector Graphic (SVG). 
An SVG is composed of many vector primitives, 
\yr{which can be SVG paths, ellipses, circles, or rectangles, etc.}
Following previous works~\cite{diffvg,im2vec,live}, we employ \yr{the} SVG paths \yr{as the shape primitive, where each SVG path} defines a region constructed by multiple cubic Bézier curves connected end-to-end with certain color. 
With the parameters of these \yr{SVG} paths, the rasterized image can be rendered in any resolution. 
To obtain the path parameters, some previous methods~\cite{diffvg,live} optimize the path parameters directly to minimize the distance between the input image and the rendered image, which achieves good reconstruction quality but \yr{requires} long optimization time. 
To speed up the vectorization process, some deep-learning-based methods~\cite{im2vec} employ a deep-learning model to predict the SVG path parameters, but 
\yr{struggle to vectorize complex images.}}

\hut{To achieve good vectorization quality \yr{with} high efficiency, we propose SuperSVG, a deep-learning-based image vectorization method that translates \yr{images} into \yr{SVG} path parameters automatically. 
To improve the model ability \yl{to vectorize} complex images, we segment the input image into different parts, within which the pixels share similar colors \yr{and} textures, 
and then 
\yr{vectorize}
each part separately, where superpixels are used for image segmentation \yl{as they tend to maintain compactness, uniformity and regularity, particularly suitable for our task}.}
For each superpixel \yl{$x \in \mathcal{X}$ where $\mathcal{X}$ is the set of all superpixels}, our model converts it into a sequence of path parameters, where each path is composed of several cubic B\'ezier curves \yr{and has a fill color}, with \yr{a} total \yr{of} $N_p$ parameters. 
With the predicted path parameters for each superpixel, we employ the differentiable renderer $R(\cdot)$ from DiffVG~\cite{diffvg} to get the rendered image $\hat{I}$ in pixel space, which is \yr{expected to be} close to the input image $I$. 

\yr{We propose a two-stage self-teaching framework,}
composed of a coarse-stage model $E_c$ to reconstruct the main structure and a \yl{refinement} model $E_r$ to enrich the details, \yr{where the predicted paths from coarse-stage model are used to guide the refinement model in vectorization}. 
$E_c$ \yr{takes} the 
superpixel $x$ \yr{as input} and \yl{outputs} $n$ paths $S=\{s_1,s_2,\cdots s_n\}$ \yr{to reconstruct main structure};
while $E_r$ \yr{takes} both the rendered image $R(S)$ and target superpixel $x$ \yr{as inputs}, and outputs $m$ paths $S'=\{s'_1,s'_2 \cdots s'_m\}$ to refine \yr{details}. 
Combining all the predicted $S$ and $S'$ for each superpixel produces the final SVG result.

\subsection{\hut{Superpixel-based Coarse Reconstruction}}

{\bf Superpixel decomposition.} \hut{
\yr{Considering} the optimization-based methods suffer from a long optimization time, our SuperSVG builds upon neural networks \yl{to efficiently predict SVG paths.}} 
\yr{However, as neural networks have difficulties in directly vectorizing
complex images~\cite{dziuba2023image}}
we \yr{therefore} simplify the 
\yr{task}
to 
\yr{vectorizing}
a certain part of the image containing homogeneous colors and textures. 
Since superpixel algorithms provide a good tool to decompose \yl{images} based on local pixel color and also ensure alignment of the regions with the image boundaries, we segment the input image into superpixels, and our model reconstructs each superpixel with scalable vectors. \yl{Superpixels also tend to be more regular, making them easier for vectorization.}
Specifically, we utilize SLIC~\cite{slic} to decompose the input image \yr{into superpixels}. 
\yr{We set the compactness parameter as $30$ for SLIC to make superpixels more regular.}

{\bf Coarse-stage model.}
\hut{For a superpixel $x$ with mask $\yl{mask}$ \yl{(indicating those pixels within $x$ with 1 and 0 otherwise)}, 
we first 
\yr{design}
the coarse-stage model $E_c$ to vectorize the main structure by predicting the \yr{SVG} path sequence $E_c(x)=S=\{s_1,s_2,\cdots s_n\}$.} 
Inspired by AttnPainter~\cite{attnpainter}, $E_c$ is composed of a Vision Transformer \yl{(ViT)} encoder~\cite{dosovitskiy2020vit} and a cross-attention module followed by a self-attention module, which is shown in Fig.~\ref{fig:main framework}.

\hut{Specifically, the ViT encoder first encodes the input superpixel $x$ into image feature $T_f$.
To control the number of output paths \yr{($n$)} and the parameter number in each path \yr{($N_p$)}, we employ a cross-attention module \yr{to} calculate the correlation between the image feature $T_f$ and 
\yr{$n$ $N_p$}-dimensional learnable path queries $T_l$, \yr{and output an} intermediate feature $T'_f$ \yr{with} the shape of path parameters 
\yr{($n\times N_p$)}, which is formulated as:
}
\begin{equation}
\begin{aligned}
    T'_f=Softmax(\frac{(W_Q T_l)(W_K T_f)^T}{\sqrt{d}}) (W_V T_f),
\end{aligned}
\end{equation}
\yl{where $W_Q$, $W_K$ and $W_V$ are learnable query, key and value matrices.}

\hut{Then, a self-attention module is employed to process the image feature $T'_f$ and project it into the parameter space with $n$ paths, where each path contains $N_p$ parameters. }


\begin{figure}[t]
\centering
\includegraphics[width=0.4\textwidth]{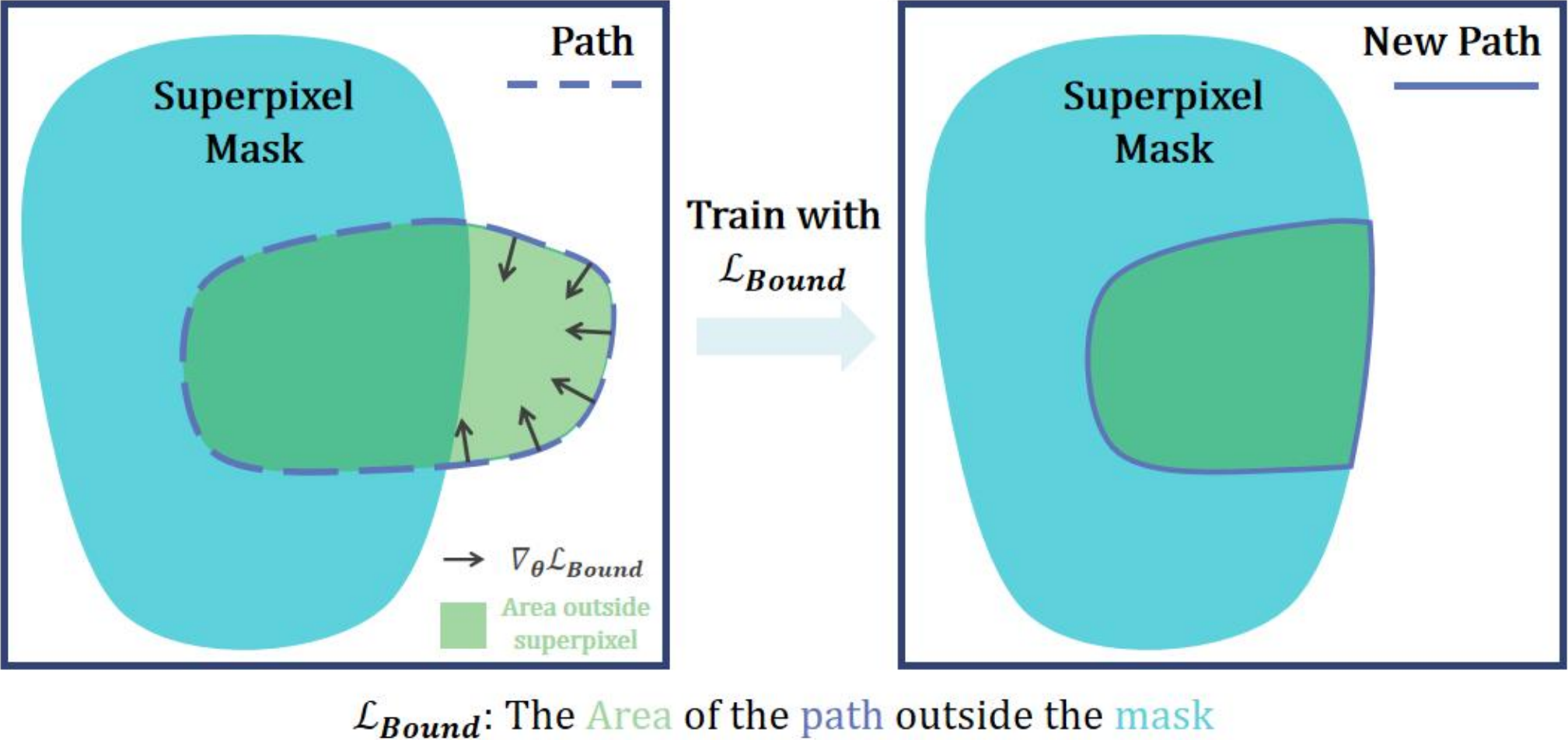}
\vspace{-0.1in}
\caption{Illustration of our \yl{boundary} loss $\mathcal{L}_{Bound}$, which computes the area of the SVG paths that are outside the superpixel mask, \yr{and guides the paths to be inside the superpixel}.}
\label{fig:boundary loss}
\vspace{-0.15in}
\end{figure}

{\bf Training objectives.} {\it 1) Normalized Reconstruction Loss:}
We employ the differentiable renderer $R(\cdot)$ in DiffVG~\cite{diffvg} to render \yr{a raster image} $\hat{x}=R(S)$ from the predicted path parameters $S$. 
Then, we train the coarse-stage model $E_c$ by minimizing the normalized \yr{$\mathcal{L}_2$} distance between the \yr{rendering} $\hat{x}=R(S)$ and the target image:
\begin{equation}
    \begin{aligned}
        \mathcal{L}_2= \|\hat{x}\yl{-}x\|^2 \cdot \frac{\sum_\yl{p} \yl{mask(p)}}{w h},
    \end{aligned}
\end{equation}
where $w$ and $h$ are the width and height of the superpixel image $x$ and the superpixel mask \yl{$mask$}, and $mask(p)$ indicates the mask value (1 or 0) for pixel $p$.

{\it 2) Boundary Loss:}
To avoid the \yr{SVG} path from crossing the superpixel boundary, we \yr{propose} boundary loss to \yr{guide} the paths to be inside the superpixel. 
We set the color of \yr{all predicted SVG} paths \yr{to} 1 \yr{(white)} to get a new path sequence $S_{binary}$. 
Then, we compute the boundary loss by:
\begin{equation}
    \begin{aligned}
        \mathcal{L}_{Bound}= \mathop{\mathbb{E}}\limits_{\crr{p\sim mask}} (R(S_{binary})\cdot (1-\yl{mask})),
    \end{aligned}
\end{equation}
\yr{Since $(1-\yl{mask})$ is 1 outside the superpixel mask and 0 inside the mask, the loss term calculates the area of the paths that are outside the superpixel. When some SVG paths cross the superpixel boundary, the path area outside is penalized; while}
when all \yr{SVG} paths are inside the superpixel, $\mathcal{L}_{Bound}$ reaches 0 \yr{(Fig.~\ref{fig:boundary loss})}.

{\it 3) Path Efficiency Loss:}
\hut{To enable our model to reconstruct the maximum amount of information with the fewest paths, we \yr{propose the} path efficiency loss $\mathcal{L}_{PE}$.} 
Specifically, for each path $i$, an additional opacity parameter $\beta_i$ is predicted. 
\yl{We treat the path as visible 
if 
\yr{$\beta_i \geq 0.5$},
and the loss $L_{PE}$ penalizes the case with more 
\yr{visible} paths, \yr{{\it i.e.,} to encourage reconstructing the image with as few paths as possible}, calculated as}:
\begin{equation}
    \begin{aligned}
        \mathcal{L}_{PE}= \textstyle \sum_{i=1}^n Sign(\beta_i-0.5),
    \end{aligned}
\end{equation}
\yr{where $Sign(\cdot)$ is the sign function.} Since $Sign(\beta_i)$ is not differentiable, we approximate $\frac{\partial Sign(\beta_i)}{\partial \beta_i}$ by $Sig(\beta_i)(1-Sig(\beta_i))$, where $Sig(\cdot)$ is the sigmoid function.

The final training objective is formulated as:
\begin{equation}
    \begin{aligned}
        E_c^*=
        \mathop{\arg\min}\limits_{E_c} \mathcal{L}_2+\lambda_{Bound} \mathcal{L}_{Bound} + \lambda_{PE} \mathcal{L}_{PE}.
    \end{aligned}
\end{equation}

\begin{figure}[t]
\centering
\includegraphics[width=0.36\textwidth]{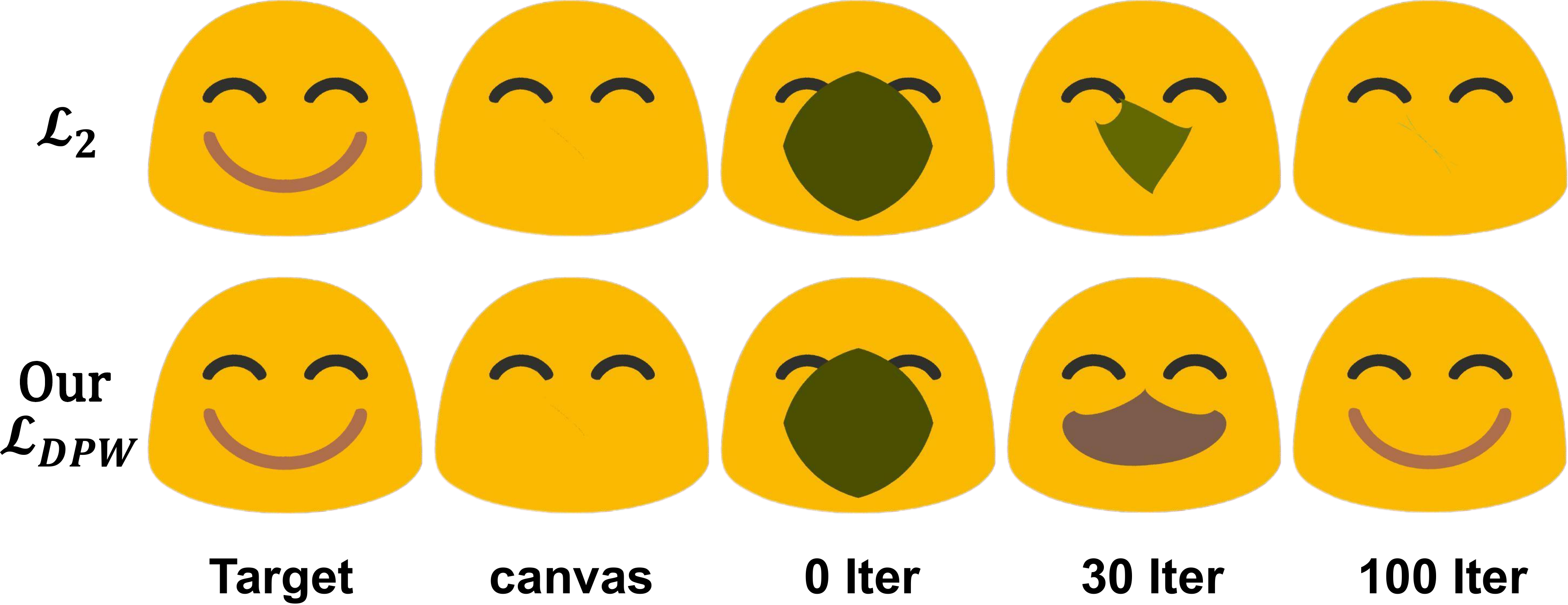}
\vspace{-0.1in}
\caption{Problem of training the \yr{refinement} model \yr{with $\mathcal{L}_2$ loss alone}: 
optimizing a newly-added path on the canvas \yr{by $\mathcal{L}_2$} gradually pulls it to disappear \yl{(as a suboptimal local minimum)}. \yr{With our proposed coarse paths guided training and DPW loss, the added path is successfully optimized to resemble the target.}
}
\label{fig:problem_of_model2}
\vspace{-0.15in}
\end{figure}

\subsection{
\hut{\yl{Coarse Paths} Guided \yr{Refinement} Stage}}
\label{sec:refinement stage}
In the coarse reconstruction stage, our coarse-stage model $E_c$ can \yr{output \yl{an} SVG that} captures and reconstructs the main structure of the input superpixel $x$. 
The \yr{rendered} image \yr{from the SVG} (denoted as $c_1$) resembles $x$ in general, but 
lacks some image details, especially when the superpixel is complex. 
To enrich image details, we employ a \yr{refinement} model $E_r$ to \yr{predict more SVG paths to add} 
more details based on the current canvas $c_1$. 

{\bf Model framework.} 
Different from the coarse-stage model $E_c$ that only \yr{takes} the 
\yr{target superpixel $x$}
\yr{as input}, 
the \yr{refinement} model $E_r$ \yr{takes} both the current canvas $c_1$ \yr{(rendered from the coarse stage output)} and the target 
\yr{superpixel}
$x$ \yr{as inputs}, 
and predict new paths $S'=\{s'_1, s'_2 \cdots s'_m\}$ 
\yl{to be overlaid}
onto the canvas \yr{to refine details}. 
Specifically, 
3 convolution layers followed by ReLU \yl{activations} are employed to fuse the \yr{two input} images into a feature map. 
After getting the fused feature map, $E_r$ shares the same structure as the coarse\yr{-}stage model $E_c$, which encodes the fused feature map by a ViT Encoder and maps the encoded features into path parameters by a cross-attention and a self-attention layer. 
To accelerate the training process, we inherit the \yl{weights} of the ViT encoder in $E_c$ as an initialization. 

{\bf Local optimal solution with $\mathcal{L}_2$ loss.}
The goal of the \yr{refinement} model is to reconstruct more details of the input superpixel based on the current canvas. 
\yr{A simple $\mathcal{L}_2$ loss defined as follows \yl{is} used as the reconstruction loss:}
\begin{equation}
    \begin{aligned}
        \mathcal{L}_2=\|
        \yr{x}-R([E_c(x),E_r(x,c_1)])\|^2\cdot \frac{\sum_p mask(p)}{w h},
    \end{aligned}
    \label{eq:L2 loss of model 2}
\end{equation}
\yr{where the predicted path sequences by coarse-stage model $E_c$ and \yr{refinement} model $E_r$ are concatenated together to get the final SVG result, and the rendered image of which is expected to resemble the input 
superpixel $x$.}

\yr{However,} 
the \yr{refinement} model $E_r$ trained \yr{with} Eq.(\ref{eq:L2 loss of model 2}) \yr{alone} tends to predict paths that are extremely small in area, \yr{or} even 
invisible. 
\yr{In Fig.~\ref{fig:problem_of_model2}, we use an example to illustrate this phenomenon more clearly: we newly add a path onto the canvas and optimize the path parameters with $\mathcal{L}_2$ loss;} 
\hut{it can be seen that the new path gradually shrinks and finally disappears in the canvas.} 
\yr{A possible reason is that the coarse stage result} $c_1$ is already close to $x$, and a local optimal solution for $E_r$ is to overlap nothing \yr{onto $c_1$, which is better than adding a sub-optimal path and can} prevent the $\mathcal{L}_2$ distance from increasing. 

{\bf \hut{Coarse \yl{paths} guided training framework.}}
To avoid the \yr{refinement} model from falling into \yl{poor} local optimum, we propose a \hut{coarse \yl{paths} guided} framework, which inherits the knowledge from the coarse-stage model to help train the \yr{refinement} model with an additional constraint on the \yr{SVG path} parameters. 
\yr{As illustrated in Fig.~\ref{fig:main framework},} for an input superpixel $x$, we first \yr{use the coarse-stage model to} predict \yr{a 
\yl{coarse level}} path sequence $S=\{s_1,s_2\cdots s_n\}$. 
Then, we randomly choose a value $k\in(1,n-m)$ and split the \yr{predicted} path sequence into two subsequences: $S_1=\{s_1, s_2, \cdots, s_k\}$ and $S_2=\{s_{k+1},s_{k+2},\cdots, s_n\}$. 
\yr{The subsequence $S_1$ is then rendered into $c_1=R(S_1)$ and used as the input canvas for the 
\yr{refinement} model $E_r$, while the remaining subsequence $S_2$ can be regarded as a pseudo ground truth for the output path sequence of $E_r$.
\yl{Specifically,} when training $E_r$, in addition to the previous constraints that operate in the pixel space, we add a new constraint on the {\it path parameter space}, which minimizes the distance between path sequences $S'$ and $S_2$.}

\begin{algorithm}[t]
	\renewcommand{\algorithmicrequire}{\textbf{Input:}}
	\renewcommand{\algorithmicensure}{\textbf{Output:}}
	\caption{\small{Forward 
 \yl{pass} to \yl{efficiently} compute $\mathbf{dpw_\gamma(\yr{S},\yr{S'})}$.}}
	\label{alg1}
	\begin{algorithmic}[1]
 \small
		\REQUIRE $\yr{S}$, $\yr{S'}$, smoothing $\gamma \ge 0$, distance function $d$
		\STATE  $p_{0,j} = 0; p_{i,0} = q_{i,0} = q_{0,j} = \infty, i \in [\![n]\!], j \in [\![m]\!]$  
		\FOR {$j = 1, ..., m$}
            \FOR {$i = 1, ..., n$}
                \STATE $ p_{i,j} = d_{i,j} + \min^{\gamma}(q_{i,j-1},p_{i,j-1}) $
                \STATE $ q_{i,j} = \min^{\gamma}(q_{i-1,j},p_{i-1,j}) $
            \ENDFOR
        \ENDFOR
		\ENSURE $(\min^{\gamma}(p_{n,m},q_{n,m}),P,Q)$  
	\end{algorithmic}  
\label{alg:dpw}
\end{algorithm}


\begin{figure}[t]
\centering
\includegraphics[width=0.42\textwidth]{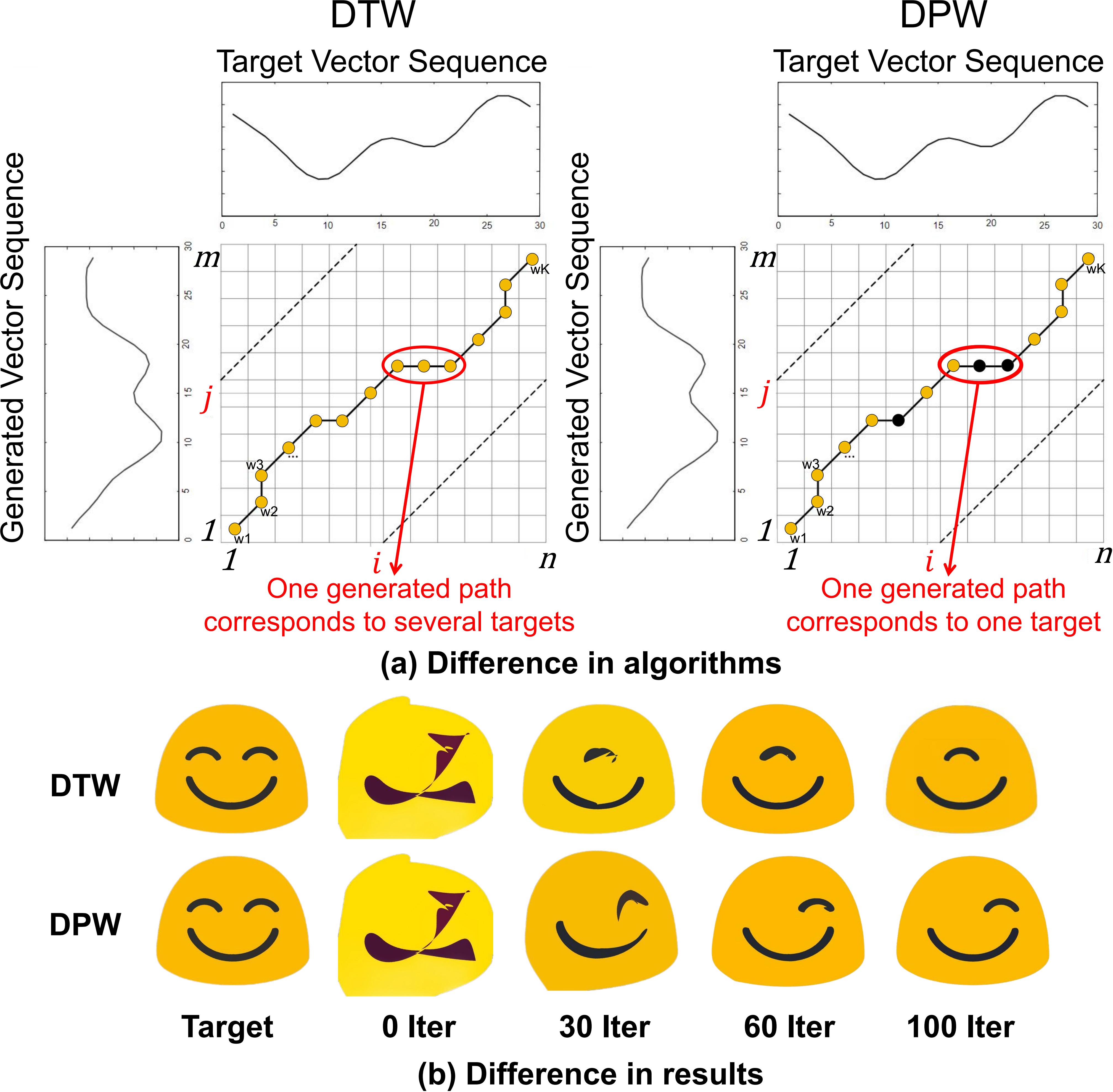}
\vspace{-0.05in}
\caption{
\yr{D}ifference between DTW and \yr{our} DPW. 
\yr{a)} Both the 
DTW and DPW loss \yr{calculate} the sum of distances of elements colored yellow. 
The difference is that one generated \yr{path} can only match one target \yr{path} in DPW to avoid averaging \yr{several target paths}. 
\yr{b)} The comparison between the training processes. 
}
\label{fig:dpw_dtw}
\vspace{-0.2in}
\end{figure}

In detail, \yr{during training,} the \yr{refinement} model $E_r$ \yr{takes} both the canvas $c_1=R(S_1)$ and the target image $x$ \yr{as inputs}, and outputs the path sequence $S'=\{s'_1, s'_2, \cdots, s'_m\}$. 
We minimize \yr{both} the distance in pixel space and \yr{path} parameter space\yr{, with the total loss formulated as}:
\begin{equation}
    \begin{aligned}
    E_r^*=\mathop{\arg\min}\limits_{E_r}\mathcal{L}_{\yr{2}} + \lambda_{DPW} \mathcal{L}_{DPW} + \lambda_{Bound} \mathcal{L}_{Bound},
    \end{aligned}
\end{equation}
where $\mathcal{L}_{DPW}$ (detailed in Sec.~\ref{ssec:dpw}) is the distance between the path sequences 
\yr{$S_2$ and $S'$}
\yr{in the path parameter space}. 

\subsection{Dynamic Path Warping}
\label{ssec:dpw}

\yr{To calculate the distance between the target path sequence $S=\{s_1,s_2,\cdots s_n\}$ and the predicted path sequence $S'=\{s'_1,s'_2\cdots s'_m\}$,} \hut{Dynamic Time Warping (DTW)~\cite{softdtw} is a common\yr{ly used} metric, 
\yr{which}
finds a\yr{n optimal} matching $match$ 
\yr{between the two \hut{ordered} sequences} (the yellow points in Fig.~\ref{fig:dpw_dtw}\textcolor{red}{(a)}) to minimize the accumulated distance $\sum\limits_i^n\|s_i-s'_{\yl{match}(i)}\|^2$ where $\yl{match}(i+1)\geq \yl{match}(i)$. 
Note that the monotonicity of the matching function cannot be ignored since the path sequences are well-ordered where 
\yr{latter paths are overlaid on former paths.}
}

\hut{In our coarse \yl{paths} guided framework, we \yr{expect} the generated path sequence to be a subsequence of the target path sequence. 
However, in DTW, one generated path \yr{$s'_j$} can correspond to several target \yl{paths} \yr{$s_{i1}, ..., s_{il}$}. Therefore, when trained with DTW, one generated path \yr{tends to} become the average of several target paths. 
As shown in Fig.~\ref{fig:dpw_dtw}\textcolor{red}{(b)}, when optimizing 3 paths to match the target emoji \yr{(with 4 paths)}, one path becomes the average of the two eyebrow paths, 
\yr{which is not our desired case.}
}

\hut{To 
\yr{address this issue,}
we propose Dynamic Path Warping (DPW), where \yr{each} generated path \yr{should} match one \yr{and only one} target path, \yr{and some target paths can be skipped (to learn a subsequence),} as shown in Fig.~\ref{fig:dpw_dtw}\textcolor{red}{(a)}, 
\yr{each horizontal line only passes through one matching point (yellow).}
To compute the DPW, we define $p_{i,j}$ as the minimum accumulated distance when $s_i$ matches $s'_j$, and $q_{i,j}$ as the minimum accumulated distance when 
\yr{
$s'_j$ has been matched to one path before $s_i$ (not including $s_i$).} 
We employ dynamic programming to compute the final DPW loss $p_{n,m}$ as shown in Alg.~\ref{alg:dpw}. For each $p_{i,j}$, the distance $d_{i,j}$ between $s_i$ and $s'_j$ is added to the smaller one of $q_{i,j-1}$ and $p_{i,j-1}$. And for each $q_{i,j}$, its value takes the smaller one between $q_{i-1,j}$ and $p_{i-1,j}$ (more explanations are provided in the supplementary material). Moreover, to make Alg.~\ref{alg:dpw} differentiable, we follow SoftDTW~\cite{softdtw} to substitute the $\min(\cdot)$ operation:
}
    $$ \textstyle \min^{\gamma}(a_0, a_1, ..., a_n) = 
        \begin{cases}
        \min_{i\leq n}a_i & \gamma = 0, \\
        -\gamma \log\sum_{i=1}^{n}e^{-a_i/\gamma} & \gamma>0.
        \end{cases}
       $$
       \vspace{-0.1in}

\begin{figure*}[t]
\centering
\includegraphics[width=0.75\textwidth]{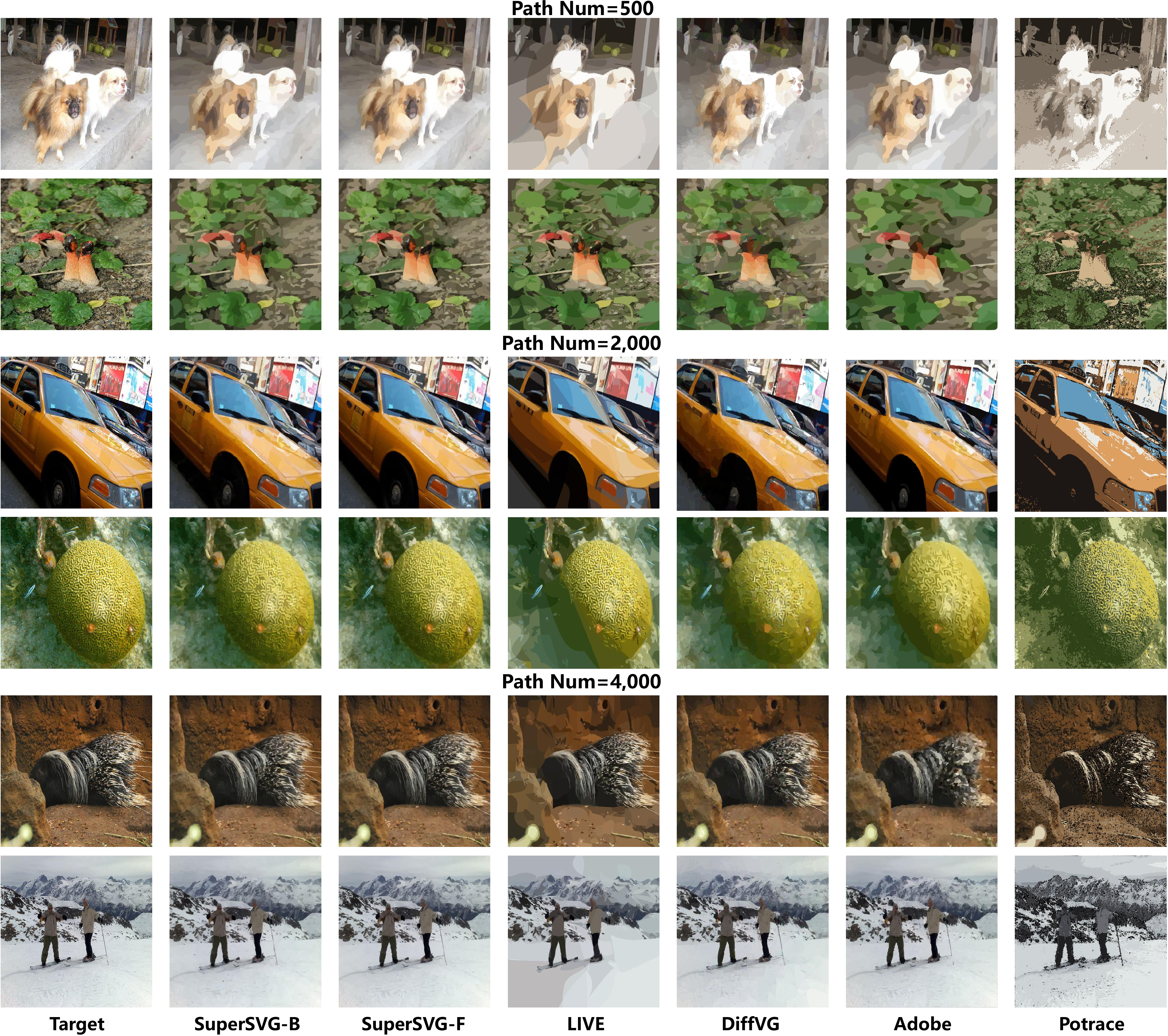}
\vspace{-0.1in}
\caption{Qualitative comparison with the state-of-the-art methods in image vectorization with different number of SVG paths.}
\label{fig:qualitative comparison}
\vspace{-0.15in}
\end{figure*}

\section{Experiments}
\subsection{Experiment Setting}

{\bf Implementation Details.} We use \yr{SVG paths composed by} cubic Bézier curves 
\yr{as the vector primitive,}
where each \yr{SVG path} is \yr{closed,} composed of 4 cubic B\'ezier curves \yr{connected end-to-end} and \yr{has a fill} color. 
Each \yr{SVG path} has \hut{28 parameters (24 for shape, 3 for color, and 1 for visibility)}. 
\yr{The coarse-stage model} is trained to predict 128 paths \yl{for each superpixel} \crr{first. Then,} \yr{the refinement model} is trained to predict 8 paths at one time. 
We train both \yr{the coarse-stage model} and \yr{refinement model} on ImageNet dataset~\cite{deng2009Imagenet}. 
We set batch size \yr{as} 64 and learning rate \yr{as} $2.5\times 10^{-4}$. 
We train \yr{the coarse-stage model} for 200K iterations with 5K warm up iterations, and train \yr{the refinement model} for 200K iterations with $\lambda_{DPW}$ decreasing from $1\times 10^{-3}$ to $0$ in 10K iterations \yr{uniformly}. 
\hut{In the following experiments, we implement our model \yr{with two versions}: 
\textbf{1)} \textbf{SuperSVG-B,} that decomposes the image into superpixels and vectorizes them by the coarse-stage model and refinement model, 
and \textbf{2) SuperSVG-F,} which finetunes the \yr{SVG parameters} from SuperSVG-B with $\mathcal{L}_\yl{2}$ loss, \yl{which takes around 10 seconds for optimization}. \hut{All experiments are carried out on an NVIDIA GeForce RTX 4090 24GB GPU.}
}

{\bf \yr{Evaluation} Details.} 
\yr{For quantitative evaluation and comparison, we test} our model on 
1,000 images \yr{randomly selected from ImageNet test set, and convert each image} into \yr{SVGs with} 500, 2,000 and 4,000 paths respectively. 
With the predicted SVG paths, we evaluate the reconstruction \yr{accuracy of the output SVG} with the following 4 metrics: \textit{\textbf{1)} MSE Distance} and \textit{\textbf{2)} PSNR} to measure the pixel distance between the \hut{input image and \yr{the rendering from} SVG}; \textit{\textbf{3)} LPIPS}~\cite{zhang2018lpips} to evaluate the perceptual distance, and \textit{\textbf{4)} SSIM}~\cite{ssim} to measure the structural distance.
\yr{We further compare with Im2Vec~\cite{im2vec} on EMOJIS dataset~\cite{notoEmoji}.}



    

\subsection{Image to 
\yr{SVG Comparison}}

{\bf State-of-the-art methods.} The state-of-the-art methods can be classified into 3 categories: 
1) Algorithm-based methods: 
{\it Potrace}~\cite{potrace} employs edge tracing to vectorize binary images. 
\yr{To process color images, we follow Color Trace\footnote{\url{https://github.com/HaujetZhao/color-trace}} to first quantize color images into different layers and then convert each layer to SVG using Potrace.}
{\it Adobe Illustrator}~\cite{AdobeIllustrator} is a widely-used commercial software which convert\yr{s an} image into SVG through image tracing.
2) Deep-learning-based methods: {\it Im2Vec}~\cite{im2vec} encodes the target image into latent and predicts the vector paths with LSTM \crr{(\#suppl.)}; 
and 3) Optimization-based methods: {\it DiffVG}~\cite{diffvg} optimizes path parameters with random initialization and {\it LIVE}~\cite{live} employs layer-wise optimization to ensure a good vectorization structure.
\yr{We use the official codes of these methods and default settings for comparison.}

{\bf \yr{Qualitative} Comparison on ImageNet.}
We compare with the \yr{state-of-the-art vectorization} methods in reconstruction accuracy \yr{on ImageNet}. 
Specifically, we conduct the comparison experiments under path numbers 500, 2,000 and 4,000\footnote{\hut{\yr{Since} Adobe and Potrace \yr{outputs have different number of parameters in each path}, we \yr{keep} their \yl{output} parameter number the same as ours}.}. 
The qualitative results are shown in Fig.~\ref{fig:qualitative comparison}.
\hut{It can be seen that Potrace cannot reconstruct the image well. 
LIVE loses a \yr{lot} of details in relatively smooth areas due to its emphasis on regions with substantial color variations. 
DiffVG and Adobe work better when the path number increases, but they reconstruct fewer details than our SuperSVG.}
In comparison, our SuperSVG\yr{-B} reconstructs most of the details 
with a short inference time. 
\yr{And by optimizing the SVG parameters from SuperSVG-B with only 10 seconds, our SuperSVG-F achieves the best reconstruction accuracy under different SVG path numbers.}

{\bf \yr{Quantitative} Comparison on ImageNet.}
We further conduct quantitative \yr{comparison} on 1,000 images \yr{randomly} sampled from ImageNet~\cite{deng2009Imagenet} dataset (50 images for LIVE due to the extremely \yr{long optimization time:} 
\yr{each input image takes about 6 GPU hours to optimize under 500 SVG paths).}
\yr{The quantitative results are presented in} Tab.~\ref{tab:quantitative-comparison}. \ 
Our SuperSVG achieves the best image vectorization results.


\begin{table}[t]
\small
\centering
\setlength{\abovecaptionskip}{4pt}
\setlength{\belowcaptionskip}{-0.2cm}
\setlength\tabcolsep{3pt}
\renewcommand{\arraystretch}{1.2}
\caption{Quantitative comparison between the state-of-the-arts and Ours. \hut{\textbf{Bold} and \underline{underline} for \yr{the best} and \yr{the second best} results.}}
\vspace{-0.05in}
\resizebox{1.0\linewidth}{!}{
\begin{tabular}{c|c|c|cccc}
\toprule
\#Paths & Method & Time (s) $\downarrow$& MSE $\downarrow$ & PSNR $\uparrow$ & LPIPS $\downarrow$ & SSIM $\uparrow$ \\ \midrule
\multirow{6}{*}{500} & LIVE~\cite{live} & 20,000 & \underline{0.0039} & 24.10 & 0.4467 & \underline{0.7983} \\
 & DiffVG~\cite{diffvg} & 19.29 & 0.0069 & 21.42 & 0.5319 & 0.6671 \\
 & Adobe~\cite{AdobeIllustrator} &0.87 & 0.0067 & 21.82 & 0.5595 & 0.6939  \\
 & Potrace~\cite{potrace} & 0.98 & 0.0208 & 17.85 & 0.5115 & 0.6920  \\
 & SuperSVG-B (Ours) & \textbf{0.31} & 0.0044 & \underline{24.80} & \underline{0.4452} & 0.7687 \\
 & SuperSVG-F (Ours) & 10.00 & \textbf{0.0026} & \textbf{27.46} & \textbf{0.3558} & \textbf{0.8383} \\\midrule
\multirow{6}{*}{2,000} 
&LIVE~\cite{live} & 120,000 & 0.0025 & 26.98 & 0.3994 & 0.8431 \\
&DiffVG~\cite{diffvg} &73.65 & 0.0036 & 25.88 & 0.4683 & 0.7710 \\
&Adobe~\cite{AdobeIllustrator} & 2.15 & 0.0033 & 26.23 & 0.3961 & 0.7229  \\
&Potrace~\cite{potrace} & 3.10 & 0.0160 & 19.65 & 0.4355 & 0.6997  \\
&SuperSVG-B (Ours) & \textbf{0.71} & \underline{0.0024} & \underline{27.25} & \underline{0.3648} & \underline{0.8446} \\
&SuperSVG-F (Ours) & 10.00 & \textbf{0.0017} & \textbf{29.12} & \textbf{0.2931} & \textbf{0.8828}\\ \midrule
\multirow{6}{*}{4,000} & LIVE~\cite{live} & 300,000 & 0.0024 & 26.80 & 0.3981 & 0.8446 \\
 & DiffVG~\cite{diffvg} &140.34 & 0.0025 & 27.29 & 0.3858 & 0.8492 \\
 & Adobe~\cite{AdobeIllustrator} &3.12 & 0.0035 & 25.93 & 0.3408 & 0.7297  \\
 & Potrace~\cite{potrace} & 5.11 & 0.0113 & 20.68 & 0.4380 & 0.7374 \\
 & SuperSVG-B (Ours) & \textbf{1.04} & \underline{0.0019} & \underline{28.45} & \underline{0.3187} & \underline{0.8757} \\
 & SuperSVG-F (Ours) & 10.00 & \textbf{0.0014} & \textbf{29.96} & \textbf{0.2496} & \textbf{0.9028}\\

\bottomrule
\end{tabular}}
\label{tab:quantitative-comparison}
\vspace{-0.15in}
\end{table}

\subsection{Ablation Study}

\begin{figure}[t]
\centering
\includegraphics[width=0.38\textwidth]{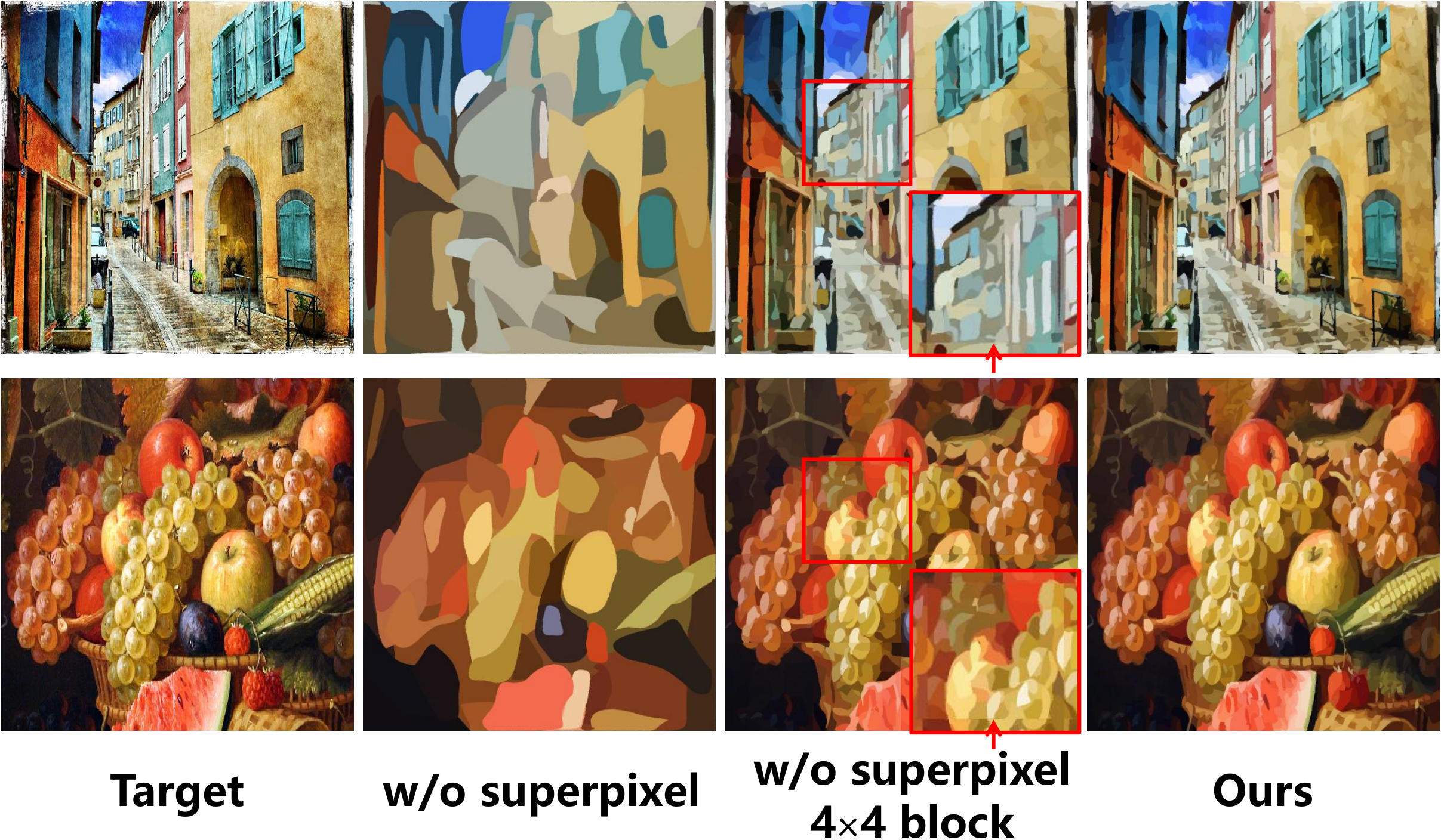}
\vspace{-0.1in}
\caption{Ablation study on the superpixel-based image vectorization framework. 
\yr{The ablated models either cannot reconstruct most details or suffer from boundary inconsistency problem.}
}
\label{fig:ablation_on_superpixel}
\vspace{-0.13in}
\end{figure}

{\bf Ablation Study on the Superpixel-based Framework.} 
We first validate the effectiveness of the superpixel-based vectorization framework. 
We train a model that directly predicts the SVG paths for an input image, \yr{without superpixel segmentation}. 
Then, we test the model in two ways: \textbf{1)} predict the SVG paths for the whole input image and 
\textbf{2)} uniformly divide the input image into $4\times 4$ blocks and \yr{vectorize} each block separately. 
We compare our model with \yr{these} two models with the same number of paths (1,000). 
The results are shown in Fig.~\ref{fig:ablation_on_superpixel} and Tab.~\ref{tab:qunatitative ablation}\textcolor{red}{(a)}. 
The model without superpixel segmentation loses \yr{many} image details. 
By introducing the block division, the model enriches the details, but the \yr{regions near} block boundaries are discontinuous (shown in red box), 
producing unnatural results.
In comparison, our superpixel-based SuperSVG-B reconstructs most details 
\yr{and} keeps the image structures well.

\begin{figure}[t]
\centering
\includegraphics[width=0.38\textwidth]{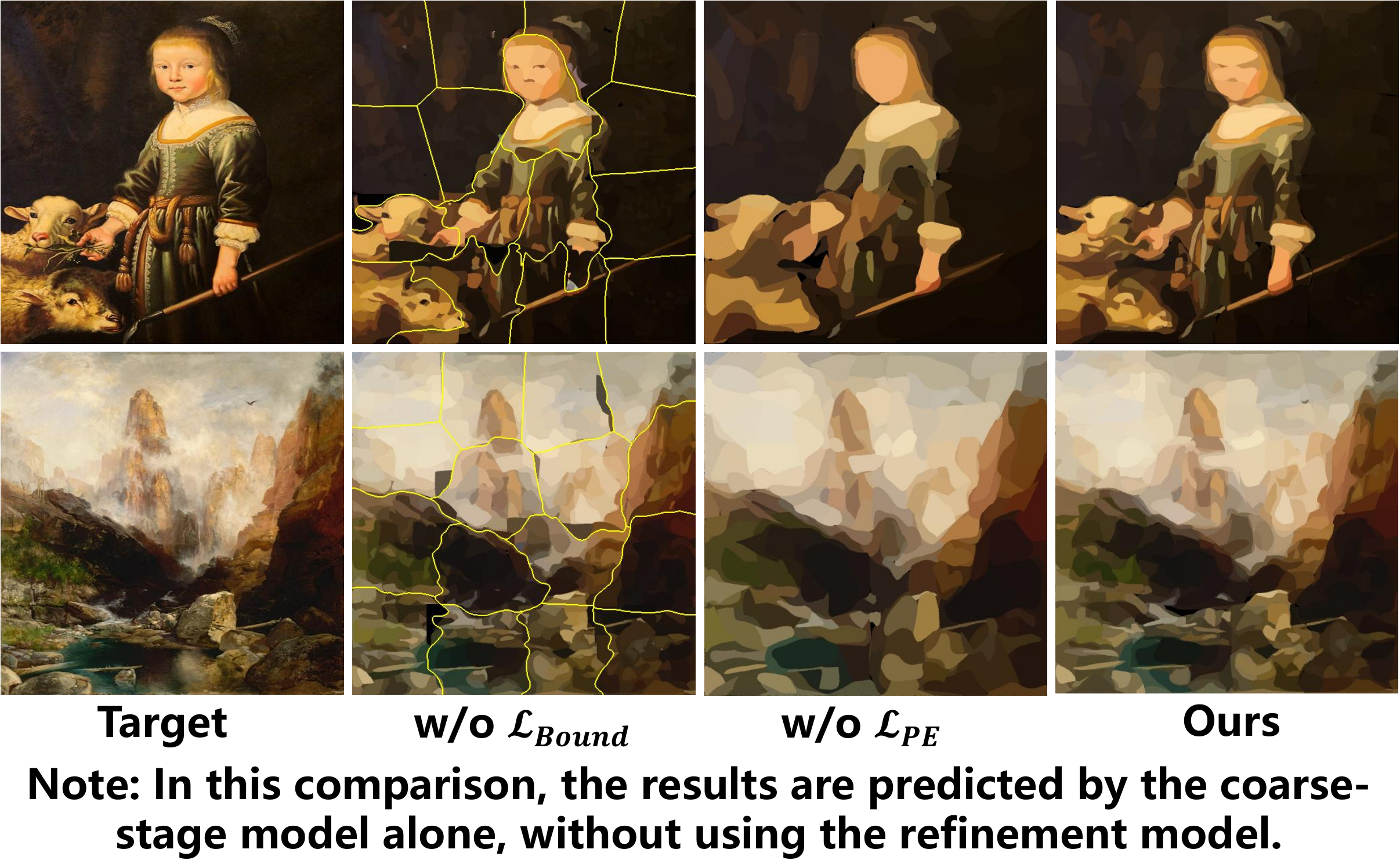}
\vspace{-0.1in}
\caption{Ablation study on the coarse-stage model. 
\yr{The ablated models either predict paths crossing superpixel boundaries or reconstruct less details than ours.}}
\vspace{-0.2in}
\label{fig:ablation_on_model1}
\end{figure}

{\bf Ablation Study on the Coarse-stage Model.} We then evaluate the effectiveness of the boundary loss $\mathcal{L}_{Bound}$ and the path efficiency loss $\mathcal{L}_{PE}$ in the coarse-stage model. We train \yr{2 ablated} models: \textbf{1)} the model without $\mathcal{L}_{Bound}$; and \textbf{2)} the model without $\mathcal{L}_{PE}$, 
and compare them with our coarse-stage model \yr{under} \textbf{500} \yr{SVG} paths.
\yr{In this comparison, we only compare vectorization using the coarse-stage model, without using the refinement model.}
\yr{The results are shown in Fig.~\ref{fig:ablation_on_model1} and Tab.~\ref{tab:qunatitative ablation}\textcolor{red}{(b)}.} \hut{The model without $L_{Bound}$ predicts some paths \yr{that cross superpixel boundaries}, resulting in worse reconstruction. 
The model without $L_{PE}$ has a poorer performance, 
which is validated \yr{by the metric results}. 
In comparison, our model outperforms the two ablated models, validating the effectiveness of the losses $\mathcal{L}_{Bound}$ and $\mathcal{L}_{PE}$ \yr{in the coarse-stage model}.}

\begin{figure}[t]
\centering
\includegraphics[width=0.38\textwidth]{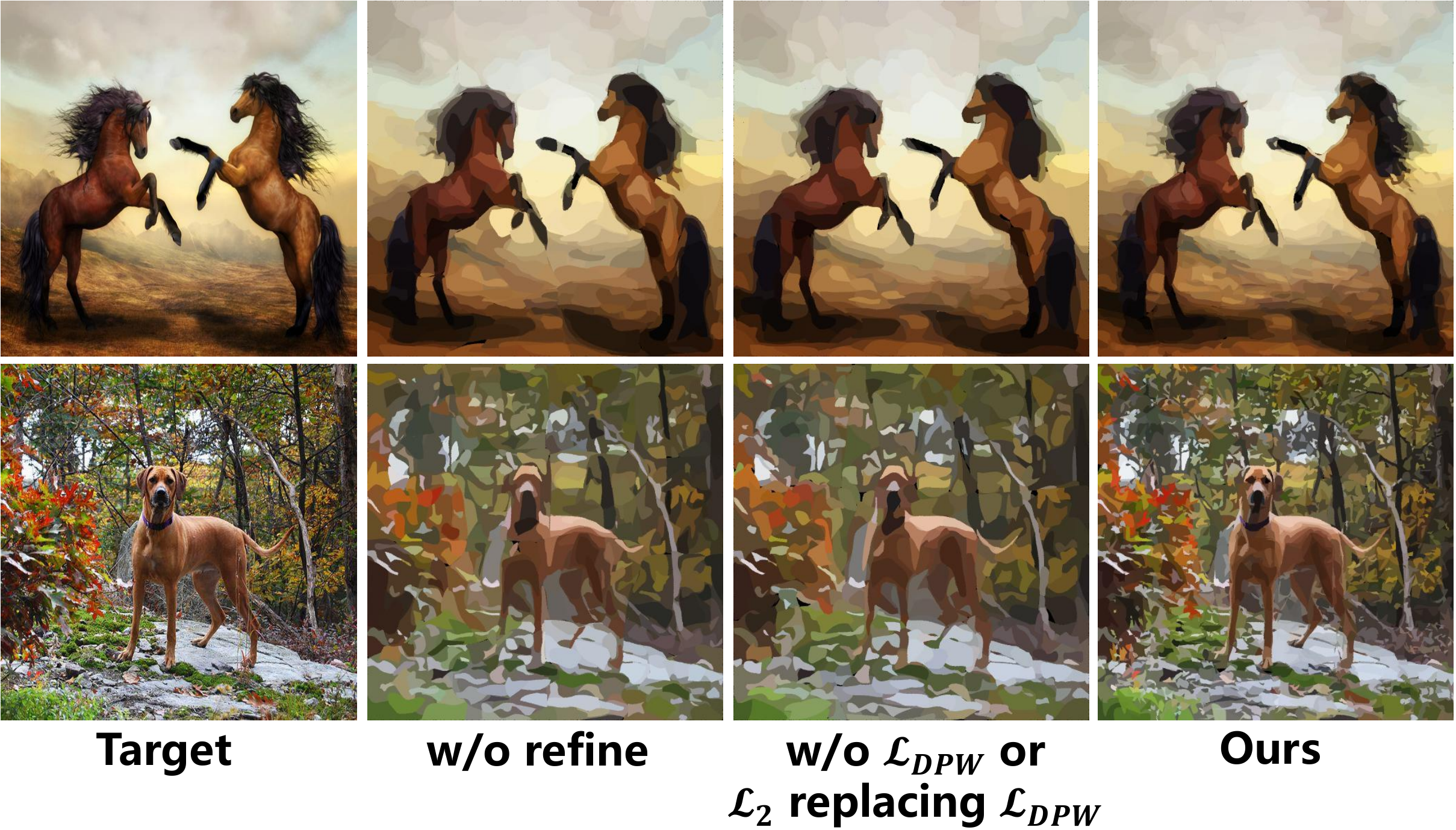}
\vspace{-0.1in}
\caption{Ablation study on 
\yr{the \yl{refinement}-stage model.}
\yr{The ablated models either lose details or converge to the poor local optimum described in Sec.~\ref{sec:refinement stage}}.
}
\label{fig:ablation_on_model2}
\vspace{-0.18in}
\end{figure}

{\bf Ablation Study on the \yl{Refinement}-stage Model.} 
Finally, we validate the effectiveness of 
\yr{the refinement stage and the DPW loss.}
We train 3 \yr{ablated} models: 
\textbf{1)} the model without the \yr{refinement} stage\footnote{\yr{Since the model without refinement does not contain refinement paths, we increase the number of coarse paths to keep the path number consistent.}};
\textbf{2)} the model without the DPW loss $\mathcal{L}_{DPW}$ \yr{({\it i.e.,} without coarse paths guidance, with pixel-wise loss only)}; and 
\textbf{3)} the model replacing DPW loss $\mathcal{L}_{DPW}$ with $\mathcal{L}_2$ loss \yr{in path parameter space}.
The results are shown in Fig.~\ref{fig:ablation_on_model2} and Tab.~\ref{tab:qunatitative ablation}\textcolor{red}{(c)}. 
The model without \yl{refinement}-stage cannot reconstruct as many details as ours. 
\yr{The ablated model without $\mathcal{L}_{DPW}$ predicts paths with a very small area or even invisible, as explained in Sec.\ref{sec:refinement stage}, thus the results look similar to the coarse-stage results.
For the ablated model replacing DPW loss $\mathcal{L}_{DPW}$ with $\mathcal{L}_2$ in path parameter space, which enforces one-to-one alignment between two paths' control points, since the constraint is too strict, the loss cannot function well in experiments, and the results look alike the results without $\mathcal{L}_{DPW}$.}
\yr{In comparison, our model outperforms the ablated models, validating the effectiveness of the refinement and DPW loss.}

\begin{table}[t]
\centering
\caption{Quantitative \yr{results of} ablation stud\yr{ies}.}
\vspace{-0.1in}
\begin{tabular}{c}
\begin{minipage}[t]{0.43\textwidth}
\centering
\resizebox{\textwidth}{!}{
\begin{tabular}{p{3.5cm}<{\centering}|cccc}
\toprule
Method & MSE $\downarrow$& PSNR $\uparrow$ & LPIPS $\downarrow$ & SSIM $\uparrow$ \\
\midrule
w/o superpixel & 0.0083 & 21.72 & 0.5057 & 0.6900 \\
$4\times4$ blocks & 0.0038 & 29.50 & 0.4299 & 0.7834 \\
\textbf{Ours} & \textbf{0.0032} & \textbf{26.04} & \textbf{0.4075} & \textbf{0.8111}\\
\bottomrule
\end{tabular}}
\caption*{(a) Ablation on superpixel-based framework.}
\vspace{-0.1in}
\end{minipage}
\\
\begin{minipage}[t]{0.43\textwidth}
\centering
\resizebox{\textwidth}{!}{
\begin{tabular}{p{3.5cm}<{\centering}|cccc}
\toprule
Method & MSE $\downarrow$& PSNR $\uparrow$ & LPIPS $\downarrow$ & SSIM $\uparrow$ \\
\midrule
w/o $\mathcal{L}_{Bound}$ & 0.0534 & 13.38 & 0.5041 & 0.6210 \\
w/o $\mathcal{L}_{PE}$ & 0.0063 & 22.98 & 0.4830 & 0.7219 \\
\textbf{Ours \yr{(Coarse)}} & \textbf{0.0045} & \textbf{24.49} & \textbf{0.4452} & \textbf{0.7673}\\
\bottomrule
\end{tabular}
}
\caption*{(b) Ablation on coarse-stage model (results are obtained by the coarse-stage model alone, without refinement model).}
\vspace{-0.1in}
\end{minipage}
\\
\begin{minipage}[t]{0.43\textwidth}
\centering
\resizebox{\textwidth}{!}{
\begin{tabular}{p{3.5cm}<{\centering}|cccc}
\toprule
Method & MSE $\downarrow$& PSNR $\uparrow$ & LPIPS $\downarrow$ & SSIM $\uparrow$ \\
\midrule
w/o refine & 0.0041 & 25.02 & 0.4375 & 0.7770 \\
w/o $\mathcal{L}_{DPW}$ & 0.0045 & 24.49 & 0.4452 & 0.7673 \\
$\mathcal{L}_2$ replacing $\mathcal{L}_{DPW}$ & 0.0045 & 24.49 & 0.4452 & 0.7673 \\
\textbf{Ours} & \textbf{0.0032} & \textbf{26.04} & \textbf{0.4075} & \textbf{0.8111}\\
\bottomrule
\end{tabular}
}
\caption*{(c) Ablation on 
\yr{\yl{refinement}-stage model.}
}
\end{minipage}

\end{tabular}

\label{tab:qunatitative ablation}
\vspace{-0.35in}
\end{table}

\section{Conclusion}
We propose SuperSVG, a novel superpixel-based vectorization model that decomposes a raster image into superpixels and then vectorizes each separately, achieving fast and accurate image vectorization.
We propose a 
\yr{two-stage self-teaching}
framework, \yr{where a coarse-stage model reconstructs main structure and a refinement model enriches details, with a novel} dynamic path warping loss that \yr{guides the refinement model by inheriting knowledge from coarse paths. Extensive experiments \hut{demonstrate} that SuperSVG achieves the state-of-the-art performance on  vectorization.}

\section*{Acknowledgments}
This work was supported by National Natural Science Foundation of China (62302297, 72192821, 62272447), Shanghai Sailing Program (22YF1420300), Young Elite Scientists Sponsorship Program by CAST (2022QNRC001), Beijing Natural Science Foundation (L222117), the Fundamental Research Funds for the Central Universities (YG2023QNB17), Shanghai Municipal Science and Technology Major Project (2021SHZDZX0102), Shanghai Science and Technology Commission  (21511101200).

{
    \small
    \bibliographystyle{ieeenat_fullname}
    \bibliography{main}
}


\clearpage

\appendix
\section{\yr{Overview}}

\yr{In this supplementary material, more details about the proposed SuperSVG method and more experimental results are provided, including:}

\begin{itemize}
    \item More details about our Dynamic Path Warping (DPW) (Section~\ref{sec:dpw});
    \item More \yl{details for the experiments} (Section~\ref{sec:more experiment details});
    \item Comparison experiments on Emoji dataset (Section ~\ref{sec:comparison on Emoji});
    \item More ablation studies (Section~\ref{sec:ablation on superpixel model});
    
    \item \yl{Additional} comparison experiments (Section~\ref{sec:more comparison experiments});
    \item More results \yl{of our method} (Section~\ref{sec:more of our results}).
\end{itemize}

\section{Details about Our Dynamic Path Warping}
\label{sec:dpw}


{\bf Problem Insight.} 
Given the generated path sequence $S'=\{s'_1,s'_2\cdots s'_m\}$ and the target path sequence $S=\{s_1,s_2\cdots s_n\}$, we aim to find the distance between the two path sequences in path parameter space. 
Denote the distance matrix between each path pair of $S$ and $S'$ as \hut{$D=\{d_{i,j}\}_{n\times m}$}, \yr{with $d_{i,j}$ \yl{being} the distance between $s_i$ and $s'_j$.} Our DPW aims to find an optimal matching function $match(j)$ that minimizes \hut{the objective function: 
\begin{equation}
\begin{aligned}
\sum_{j=1}^m \|d_{match(j),j}\|, 
\label{eq:dpw-objective}
\end{aligned}
\end{equation}
where $match(j)\geq match(j-1)$\footnote{\yr{Each generated path $s'_j$, should be matched to one and only one target path $s_i$ but a target path $s_i$ may be unmatched to any generated path $s'_j$.}}}.

\begin{figure}[t]
\centering
\includegraphics[width=0.42\textwidth]{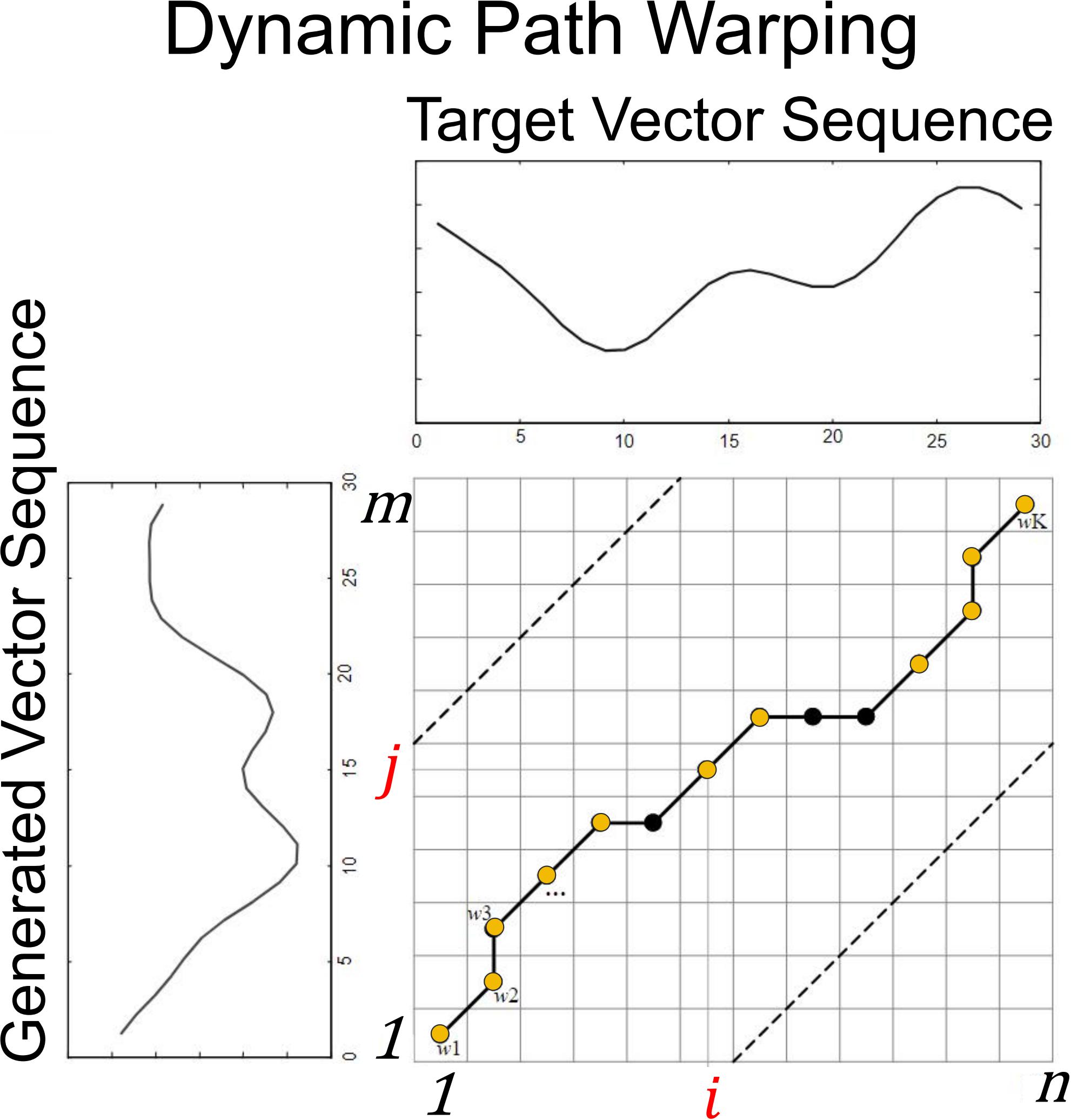}
\caption{
Illustration of our Dynamic Path Warping. 
\hut{\yr{Given a distance matrix $D_{n\times m}$ \yr{(with $d_{i,j}$ \yl{being} the distance between $s_i$ and $s'_j$)} represented as a grid, our} DPW \yr{aims to} find \yr{an optimal} \yl{grid} path composed of yellow and black points, to minimize the objective \yr{function (Eq.(\ref{eq:dpw-objective}))}. 
The yellow point $(i,j,1)$ represents \yr{a matching} $match(j)=i$, and the black point $(i,j,0)$ represents \yr{non-matching} $match(j)\neq i$ \yr{({\it i.e.,} it is not counted in the objective function)}.}
}
\label{fig:dpw}
\vspace{-0.15in}
\end{figure}

\yr{We transform the problem into finding an optimal path in a Cartesian grid.}
In Fig.~\ref{fig:dpw}, \yr{the distance matrix can be represented as an $m\times n$ grid, where} the yellow point corresponds to one mapping of $match(j)=i$, while the black point \yr{indicates non-matching ({\it i.e.,} not added in the objective function)}, 
with the yellow point denoted as $(i,j,1)$ and the black point denoted as $(i,j,0)$.
\hut{Therefore, any matching function $match(j)$, where $1\leq j\leq m$, can be represented by $m$ yellow points in the 
\yr{grid, with each yellow point in a different row.
Considering every two adjacent rows have two yellow points $(match(j-1),j-1,1)$ and $(match(j),j,1)$, with $match(j-1) \leq match(j)$,}
by adding black points between the two yellow points \yr{when they} are not adjacent, 
\yr{the $m$ yellow points and the added black points can form a path in the grid that starts from the bottom left corner to the top right corner, which only consists of rightward, upward, and diagonal up-right movements,}}
\hut{and can be further computed by dynamic programming.} 
\yr{In this way, the problem of finding an optimal matching function is transformed to finding an optimal path in the distance grid.}

\yr{Specifically,} the path should satisfy the following conditions:

1) For each row of the path, there should be one and only one yellow point, \yr{\yl{i.e.,} there will not be two adjacent yellow points in the same row (with the same $j$).}\footnote{\yr{The underlying reason is that we require one generated path to only correspond to one target path, in order to avoid a single generated path being the average of several target paths.}}

2) The path contains both the yellow points $(i,j,1)$ that \hut{represent} $match(j)=i$ and black points $(i,j,0)$ that \hut{represent} $match(j)\neq i$.

3) For each point in the path, it can only move to the right, upward, or diagonally upward to the right.

It can be easily seen that the final value of DPW only depends on the yellow points \yr{(matching points), while the \hut{values} of black points (non-matching points) are not counted in the objective function}.
Then, we can simplify the form of the path \yl{for DPW} based on the following properties.

\textbf{Property 1.}
{\it We can ignore movements that are diagonally upwards to the right \yr{(i.e., from $(i,j)$ to $(i+1,j+1)$).}}

\begin{figure}[t]
\centering
\includegraphics[width=0.42\textwidth]{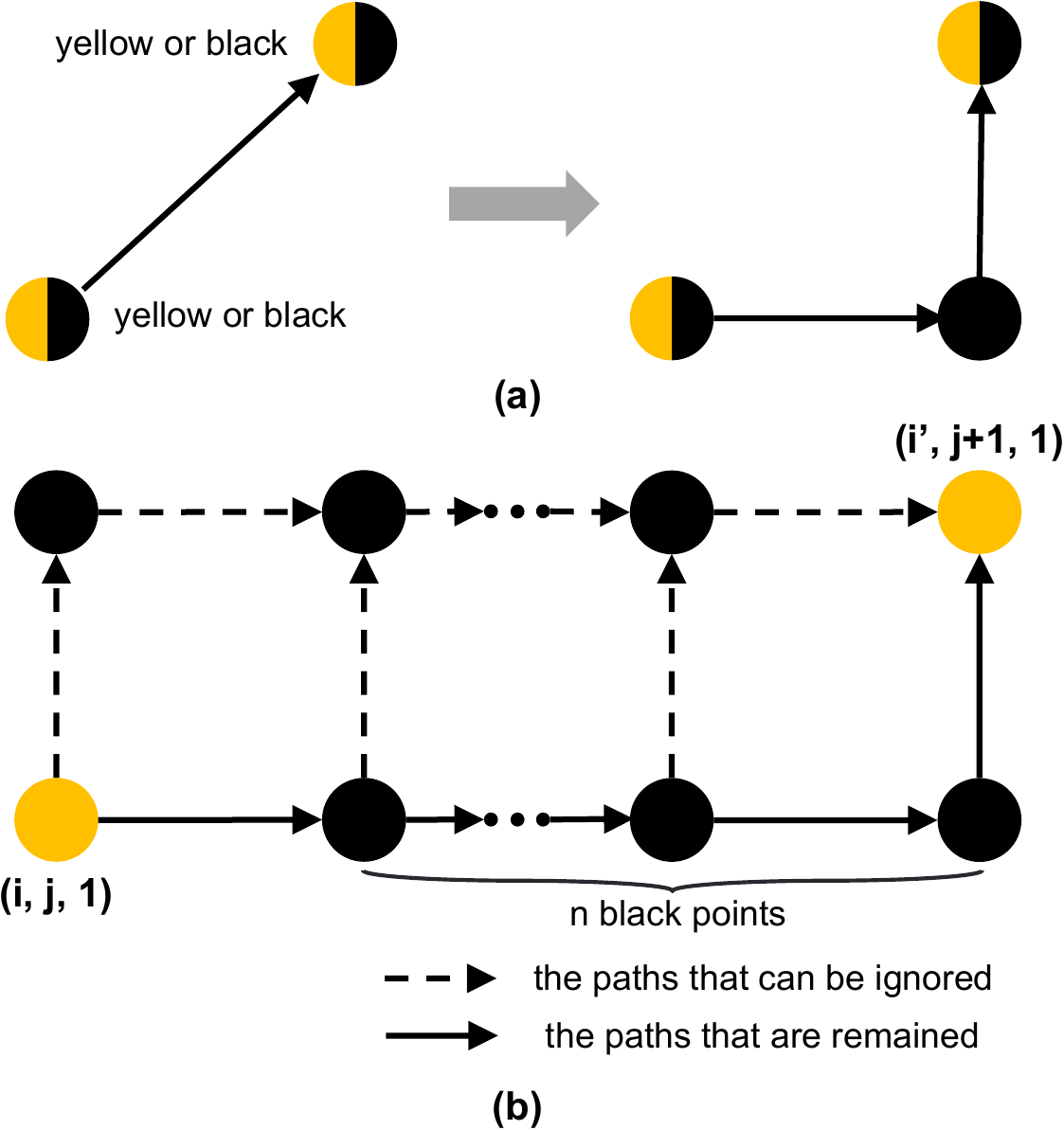}
\caption{Path simplification \yl{for DPW}. 
\hut{(a) \textbf{Property 1}: The movement from $(i,j)$ to $(i+1,j+1)$ can be simplified to moving from $(i,j)$ \yr{rightward} to $(i+1,j,0)$ \yr{(black point)} and then moving upward to $(i+1,j+1)$}. 
(b) \textbf{Property 2}: \yr{Although there are multiple paths (composed of $n$ black points) between two yellow points $(i,j,1)$ and $(i',j+1,1)$ in adjacent rows, they are equivalent in terms of their objective function values. Therefore, we can ignore} the \yr{transition} modes represented by the dashed arrows, and only need \yr{to} consider the solid arrows. 
}
\label{fig:path simplification1}
\vspace{-0.15in}
\end{figure}

\textbf{\yr{Explanation 1.}}
{\it \hut{For any \yr{diagonal up-right movement from} colored point $(i,j)$ (black or yellow) to \yr{point} $(i+1,j+1)$ (Fig.~\ref{fig:path simplification1}\textcolor{red}{(a)}), it is equal to first moving to the right \yr{to the} black point $(i+1,j,0)$ and then moving upwards to $(i+1,j+1)$\yr{:} 
By adding the intermediate black point $(i+1,j,0)$, 
1) the colors of the two endpoints $(i,j)$, and $(i+1,j+1)$ are not changed,
and 2) the intermediate black point $(i+1,j,0)$ does not contribute to the \yr{value of the DPW objective function}. 
Therefore, we can consider only rightward and upward movements.}}

\textbf{Property 2.}
{\it We can consider only four possible state transition modes: 
$(i,j,1) \rightarrow (i,j+1,1)$, 
$(i,j,1) \rightarrow (i+1,j,0)$, 
$(i,j,0) \rightarrow (i+1,j,0)$,
and $(i,j,0) \rightarrow (i,j+1,1)$.}

\textbf{\yr{Explanation 2.}}
{\it For two yellow points $(i,j,1)$ and $(i',j+1,1)$ in adjacent rows, their positional relationship can be summarized into Fig.~\ref{fig:path simplification1}\textcolor{red}{(b)}, where the number of black points~$n$ between them is larger than or equal to $0$\footnote{\yr{Since here $i=match(j)\geq match(j-1)=i'$.}}. 
When $n=0$, $(i,j,1)$ just moves upward \yr{to $(i,j+1,1)$}. 
When $n>0$, the paths between $(i,j,1)$ and $(i',j+1,1)$ can be any composition of the dashed and solid arrows that \yl{moves} from $(i,j,1)$ to $(i',j+1,1)$ in Fig.~\ref{fig:path simplification1}\textcolor{red}{(b)}, \yr{{\it i.e.,} there are different ways to add black points between the two yellow points to connect into a path}. 
However, all the \yr{possible} paths contribute to the same result since \yr{the value of the DPW objective function} only depends on the yellow points. 
\yr{Considering the different paths between these two yellow points are equal in objective function values}, we can simplify the paths by \yr{ignoring} the \yr{transition modes represented by the} dashed \yr{arrows}, 
\yr{and only need to consider the transition denoted by solid arrows.}
Therefore, there are only 4 remaining state transition modes (Fig.~\ref{fig:path simplification1}\textcolor{red}{(b)}): \\
(1) $(i,j,1) \rightarrow (i,j+1,1)$ (yellow upward to yellow), \\
(2) $(i,j,1) \rightarrow (i+1,j,0)$ (yellow rightward to black), \\
(3) $(i,j,0) \rightarrow (i+1,j,0)$ (black rightward to black), \\
(4) $(i,j,0) \rightarrow (i,j+1,1)$ (black upward to yellow).}

\begin{figure}[t]
\centering
\includegraphics[width=0.42\textwidth]{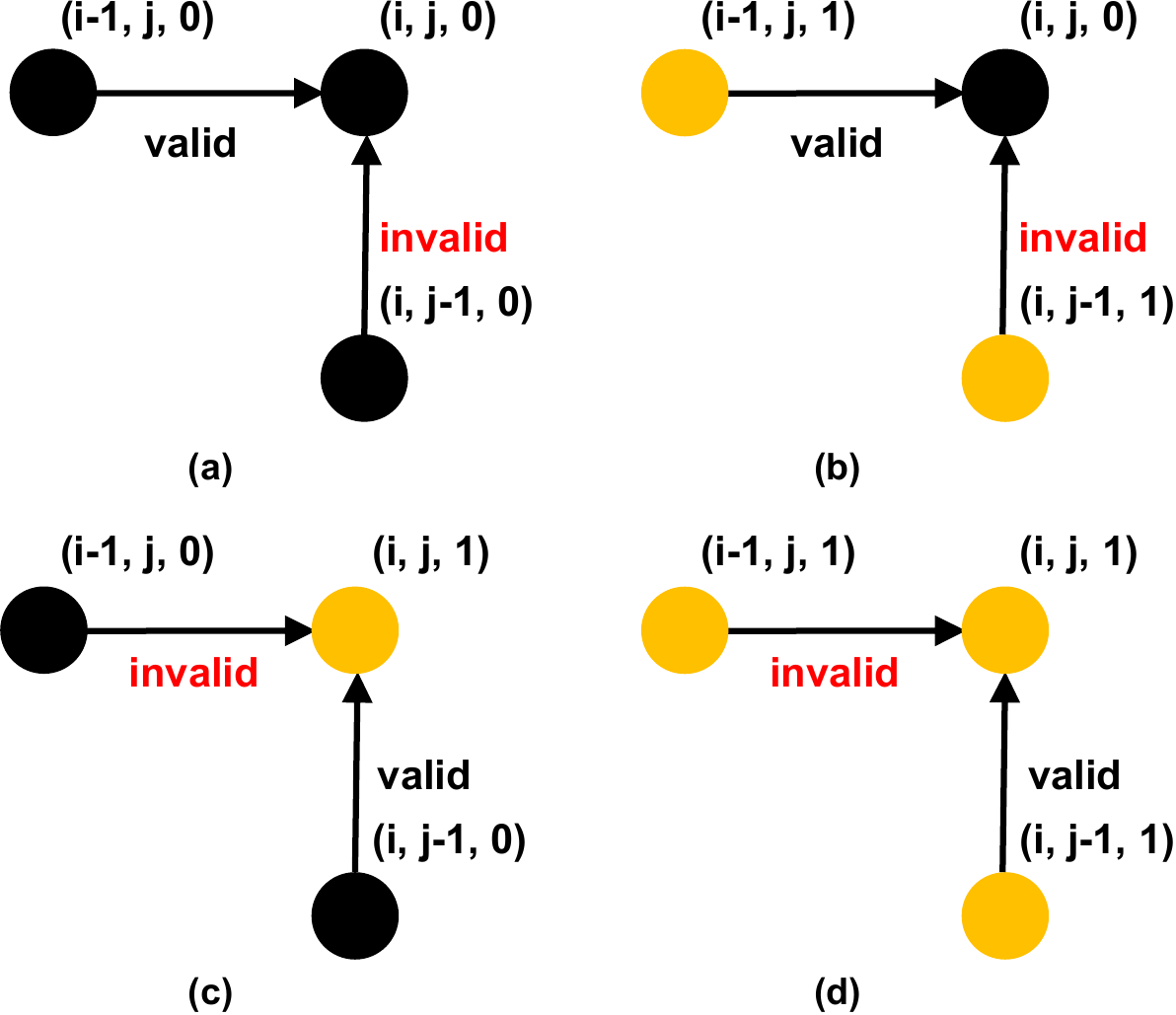}
\caption{The \yr{illustration of} valid paths \yr{and invalid paths after path simplification \yl{for DPW}}. 
\hut{There are 8 \yr{transition} modes in total. 
But according to the \textit{\yl{Properties} 1 and 2}, half of the modes can be ignored (\yr{set as} 
invalid), which greatly simplifies the dynamic programming process.}}
\label{fig:Possible paths}
\vspace{-0.15in}
\end{figure}

\textbf{Dynamic Programming Solution.} 
With the simplified state \yr{transition} mode\yr{s}, we can compute our DPW by dynamic programming. 
Specifically, we define $p_{i,j}$ as the minimum accumulated distance when going from \yr{the start point} $(1,1)$ to the \yr{yellow (matching)} point $(i,j,1)$, and $q_{i,j}$ as the minimum accumulated distance when going from \yr{the start point} $(1,1)$ to the \yr{black (non-matching)} point $(i,j,0)$. 
According to \textit{\yl{Properties} 1 and 2}, there are only 4 state \yr{transition} modes:
\begin{equation}
    \begin{aligned}
        (1) p_{i,j-1}\rightarrow p_{i,j},\\
        (2) p_{i-1,j}\rightarrow q_{i,j},\\
        (3) \yr{q}_{i-1,j}\rightarrow q_{i,j},\\
        (4) q_{i,j-1}\rightarrow p_{i,j},
    \end{aligned}
\end{equation}
which are shown in Fig.~\ref{fig:Possible paths} (yellow for $p$ and black for $q$).

Therefore, we have the state \yr{transition} function:
\begin{equation}
\begin{aligned}
 p_{i,j} &= d_{i,j} + \min^{\gamma}(q_{i,j-1},p_{i,j-1}),  \\
 q_{i,j} &= \min^{\gamma}(q_{i-1,j},p_{i-1,j}),
 \end{aligned}
\end{equation}
where $d_{i,j}$ is the distance between path \yr{$s_i$} and path \yr{$s'_j$}, and the final value of our DPW \yr{objective function} is $\min^\gamma (p_{n,m},q_{n,m})$.

\hut{Overall, our DPW can be regarded as an extended version of DTW~\cite{softdtw}, \yr{with our objective function} specifically designed for measuring the distance between two SVG path sequences. 
Although there are other versions of DTW designed for different problems, to the best of our knowledge, our DPW is the first differentiable version that aims at the specially designed objective of \yr{alignment} of SVG path sequences. }







\section{More Details for \yr{t}he Experiments}
\label{sec:more experiment details}

\textbf{More Implementation Details.} In the experiments \yr{of the main paper}, we have evaluated the performance of SuperSVG under different numbers of paths. For a certain number of paths~$n$, we assign about half of the paths to the coarse-stage model and half to the refinement-stage model.  Specifically, with our path efficiency loss $\mathcal{L}_{PE}$, our coarse-stage model predicts around $32$ visible paths for a superpixel on average. Therefore, we decompose the target image into $n_1=\frac{n}{2\times 32}$ superpixels and employ the coarse-stage model to predict SVG paths for each of them. 
\yr{We combine} all the visible paths \yr{output from the coarse-stage model, and }
employ the refinement-stage model to add more paths onto each superpixel repeatedly until the total path number reaches $n$.

\textbf{Evaluation Metrics.}
\yr{In the experiments, we use four metrics to evaluate the vectorization results, comparing the rendered image of the output SVG to the target image:}

\textbf{(1) MSE Distance:} Mean Squared Error (MSE) is a \yr{widely used} metric in image processing to assess the quality of image reconstruction. 
It measures the average squared difference between the original and reconstructed images, with lower MSE values indicating better image fidelity. 

\textbf{(2) PSNR:} The Peak Signal-to-Noise Ratio (PSNR) \yr{is} one of the most prevalent and extensively utilized metrics for assessing image quality. A higher PSNR value \yr{indicates} a superior quality of image reconstruction.  

\textbf{(3) LPIPS:} 
\yr{The} Learned Perceptual Image Patch Similarity (LPIPS)~\cite{zhang2018lpips} \yr{is} a perceptual metric utilized for assessing the similarity between two images. A \yr{lower} LPIPS value \yr{indicates} a \yr{higher} similarity between the output image and the target image.
    
\textbf{(4) SSIM:} Structure Similarity Index Measure (SSIM) ~\cite{ssim} is derived from three aspects of image similarity: luminance, contrast and structure, based on the idea that the pixels have strong inter-dependencies especially when they are spatially close. The higher \yr{the} SSIM score is, the more similar the two images are.

\textbf{Network Architecture.} 
Our \textbf{Coarse-stage model} \yr{consists of} three modules: one vision transformer encoder; one cross-attention module; and one self-attention module. 
\textbf{1)} The vision transformer encoder~\cite{dosovitskiy2020vit} 
\yr{employs the ViT implementation}
from PyTorch Image Models (timm)~\cite{timm}, which \yr{takes} an $224\times 224$ image \yr{as input} and split\yr{s} the image into patches (tokens) \yr{with} size $16\times16$. 
\textbf{2)} The cross-attention module \yr{takes} the encoded feature as the Key and Value\yr{, and takes} 
the learnable path queries as the Query. 
Then \yr{the cross-attention module} is followed by a two-layer MLP with GELU activation. 
\textbf{3)} Moreover, the self-attention module is employed to further process the output from the cross-attention layer to project the image features into path parameters. And the self-attention module is also followed by a two-layer MLP with GELU activation. 

\yr{Our}
\textbf{Refinement-stage model} 
first employ\yr{s} a three-layer convolution network with $3\times 3$ convolution kernels to \yr{encode} the current canvas and target superpixel into a $3\times 224\times 224$ feature map. 
And then it employs the same network as the coarse-stage model to project the image features into $128\times 27$-dimension output. 
At last, a fully connected layer is employed to map it into path parameters with $8\times 27$ dimension. 
\textit{
\yr{For more details, please refer to}
the code \yr{provided in the supplementary material}.}

\section{Comparison on Emoji}
\label{sec:comparison on Emoji}

\begin{figure}[t]
\centering
\includegraphics[width=0.45\textwidth]{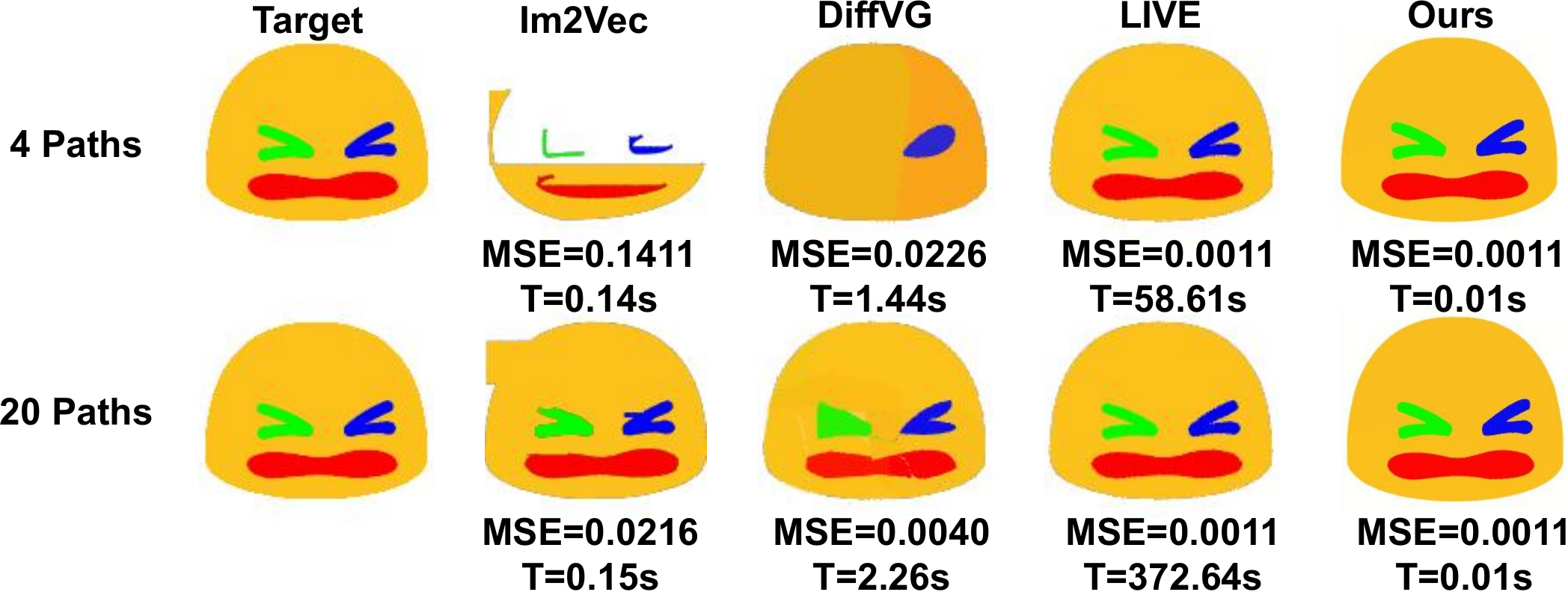}
\vspace{-0.1in}
\caption{\yr{Comparison} on EMOJIs~\cite{notoEmoji}. 
We achieve the 
\yr{best reconstruction accuracy} 
and highest speed \yr{(paths number 4 and 20)}.}
\label{fig:comparison on emoji}
\vspace{-0.2in}
\end{figure}

\yr{Since Im2Vec~\cite{im2vec} 
\yr{is domain-specific and struggles to vectorize complex images,}
we compare with it on EMOJIS dataset~\cite{notoEmoji} \yr{using its pretrained model}. 
As shown in Fig.~\ref{fig:comparison on emoji}}, our SuperSVG-B 
achieves the \yr{best reconstruction accuracy} 
and highest speed.

\section{More Ablation Studies}
\label{sec:ablation on superpixel model}

\textbf{Ablation on Different Superpixel \yr{Methods}.}
We conduct ablation studies on \yr{using different superpixel methods: we compare with using} the \yr{representative} superpixel methods including LSC~\cite{li2015lsc}, SEEDS~\cite{van2012seeds}, SpixelFCN~\cite{yang2020spixelfcn} and SNIC~\cite{snic}, as well as SLIC~\cite{slic} with different compactness (10, 20 and 30), to decompose the target image into superpixels, and then vectorize each superpixel separately. 
The comparison results with 1,000 \yr{SVG} paths are shown in Table~\ref{tab:ablation on superpixel model}.
It can be seen that SLIC 
\yr{works better with our proposed SVG synthesis framework,}
and a higher compactness in SLIC results in a better performance.

\begin{table}[t]
\small
\centering
\setlength{\abovecaptionskip}{4pt}
\setlength{\belowcaptionskip}{-0.2cm}
\setlength\tabcolsep{3pt}
\renewcommand{\arraystretch}{1.2}
\caption{Ablation \yr{study} on different superpixel \yr{methods}.}
\resizebox{0.9\linewidth}{!}{
\begin{tabular}{p{3.5cm}<{\centering}|cccc}
\toprule
Method & MSE $\downarrow$& PSNR $\uparrow$ & LPIPS $\downarrow$ & SSIM $\uparrow$ \\
\midrule
LSC~\cite{li2015lsc} & 0.0148 & 20.25 & 0.4631 & 0.6952 \\
SEEDS~\cite{van2012seeds} & 0.0042 & 24.65 & 0.4333 & 0.7777 \\
SpixelFCN~\cite{yang2020spixelfcn} & 0.0050 & 23.81 & 0.4576 & 0.7562 \\
SNIC~\cite{snic} & 0.0038 & 25.26 & 0.4125 & 0.7997 \\
SLIC-compact=10 & 0.0050 & 24.15 & 0.4388 & 0.7650 \\
SLIC-compact=20 & 0.0040 & 25.02 & 0.4205 & 0.7900 \\
SLIC-compact=30 (\textbf{Ours}) & \textbf{0.0032} & \textbf{26.04} & \textbf{0.4075} & \textbf{0.8111}\\
\bottomrule
\end{tabular}
}
\label{tab:ablation on superpixel model}
\end{table}

\textbf{Ablation on Self-attention Module.} We further conduct ablation studies on the number of self-attention modules in both our coarse-stage and refinement-stage models. 
In our model \yr{design}, we employ one self-attention module \yl{each} in coarse-stage and refinement-stage models. 
Then, we train 3 additional \yr{versions} for each of \yr{the coarse-stage and refinement-stage models} with 2, 4 and 8 self-attention modules \yr{respectively}. 
The comparison results with 1,000 \yr{SVG} paths are shown in Table~\ref{tab:ablation on self-attention module}. 
It can be seen that, as the number of the self-attention modules increases, the performance of the model does not have an obvious improvement. 
Therefore, we only employ one self-attention module \yr{to achieve} a good vectorization quality while keeping a higher efficiency and fewer \yl{learnable} parameter numbers.

\begin{table}[t]
\centering
\caption{Ablation study on the number of self-attention modules.}
\vspace{-0.05in}
\begin{tabular}{c}
\begin{minipage}[t]{0.44\textwidth}
\centering
\resizebox{\textwidth}{!}{
\begin{tabular}{p{3.5cm}<{\centering}|cccc}
\toprule
Coarse-stage Model & MSE $\downarrow$& PSNR $\uparrow$ & LPIPS $\downarrow$ & SSIM $\uparrow$ \\
\midrule
Self-attn$\times 1$ (\textbf{Ours}) & \textbf{0.0032} & \textbf{26.04} & \textbf{0.4075} & \textbf{0.8111} \\
Self-attn$\times 2$ & \textbf{0.0032} & 26.00 & 0.4079 & 0.8109 \\
Self-attn$\times 4$ & 0.0034 & 25.80 & 0.4146 & 0.8081 \\
Self-attn$\times 8$ & 0.0033 & 25.98 & 0.4080 & 0.8102 \\
\bottomrule
\end{tabular}
}
\caption*{(a) Ablation on \yr{the number of self-attention modules in} the coarse-stage model.}
\end{minipage}
\\
\begin{minipage}[t]{0.44\textwidth}
\centering
\resizebox{\textwidth}{!}{
\begin{tabular}{p{3.5cm}<{\centering}|cccc}
\toprule
Refinement-stage Model & MSE $\downarrow$& PSNR $\uparrow$ & LPIPS $\downarrow$ & SSIM $\uparrow$ \\
\midrule
Self-attn$\times 1$ (\textbf{Ours}) & \textbf{0.0032} & \textbf{26.04} & \textbf{0.4075} & \textbf{0.8111} \\
Self-attn$\times 2$ & 0.0034 & 25.78 & 0.4135 & 0.8072 \\
Self-attn$\times 4$ & 0.0033 & 25.89 & 0.4092 & 0.8096 \\
Self-attn$\times 8$ & \textbf{0.0032} & 26.01 & 0.4080 & 0.8108\\
\bottomrule
\end{tabular}
}
\caption*{(b) Ablation on \yr{the number of self-attention modules in} the refinement-stage model.}
\end{minipage}

\end{tabular}

\label{tab:ablation on self-attention module}
\vspace{-0.35in}
\end{table}

\section{Additional Comparison Experiments}
\label{sec:more comparison experiments}
\begin{figure*}[t]
\centering
\includegraphics[width=0.98\textwidth]{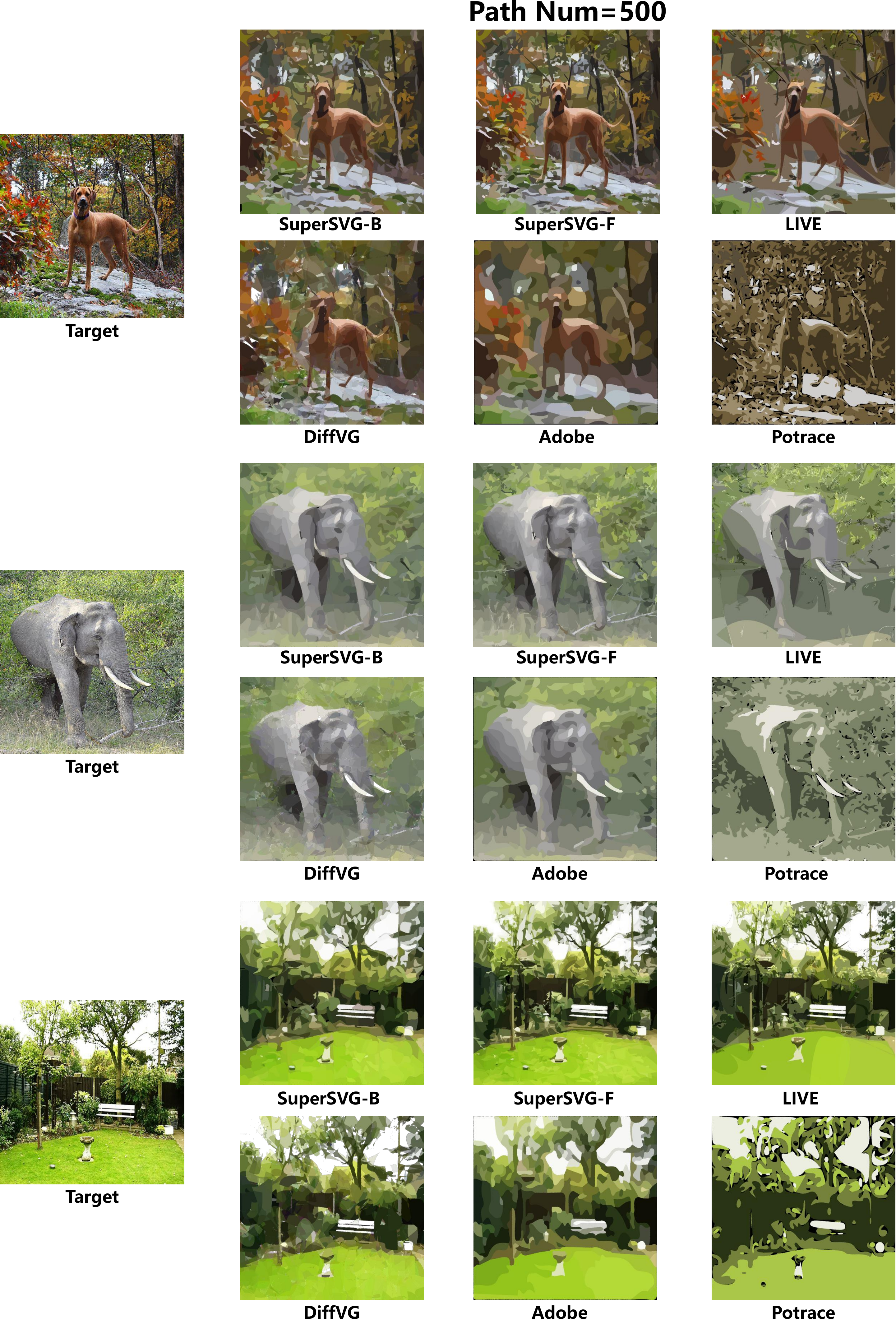}
\caption{More comparison with the state-of-the-art methods \yr{under} 500 \yr{SVG} paths.}
\label{fig:more comparison-500}
\end{figure*}

\begin{figure*}[t]
\centering
\includegraphics[width=0.98\textwidth]{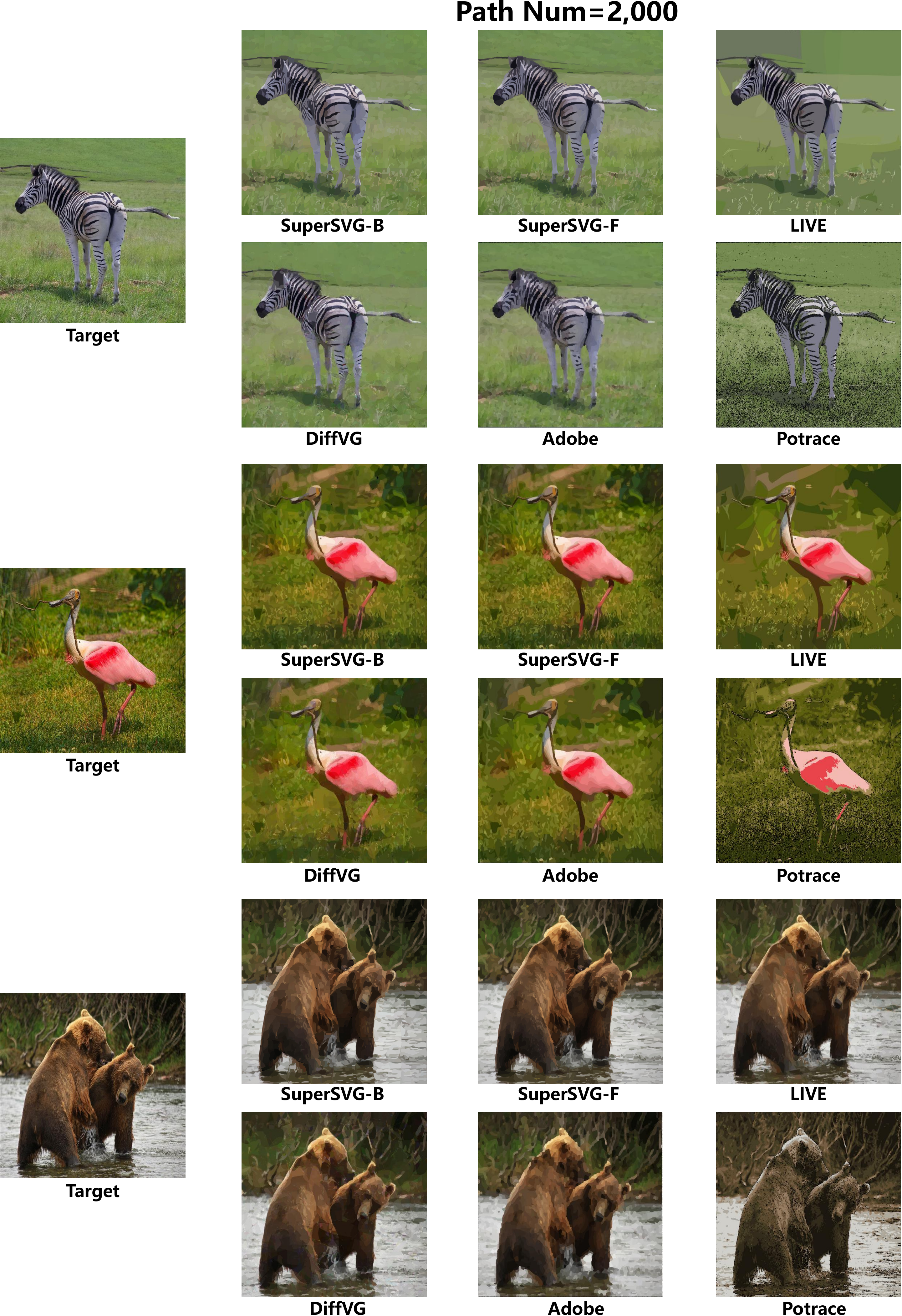}
\caption{More comparison with the state-of-the-art methods \yr{under} 2,000 \yr{SVG} paths.}
\label{fig:more comparison-2000}
\end{figure*}

\begin{figure*}[t]
\centering
\includegraphics[width=0.98\textwidth]{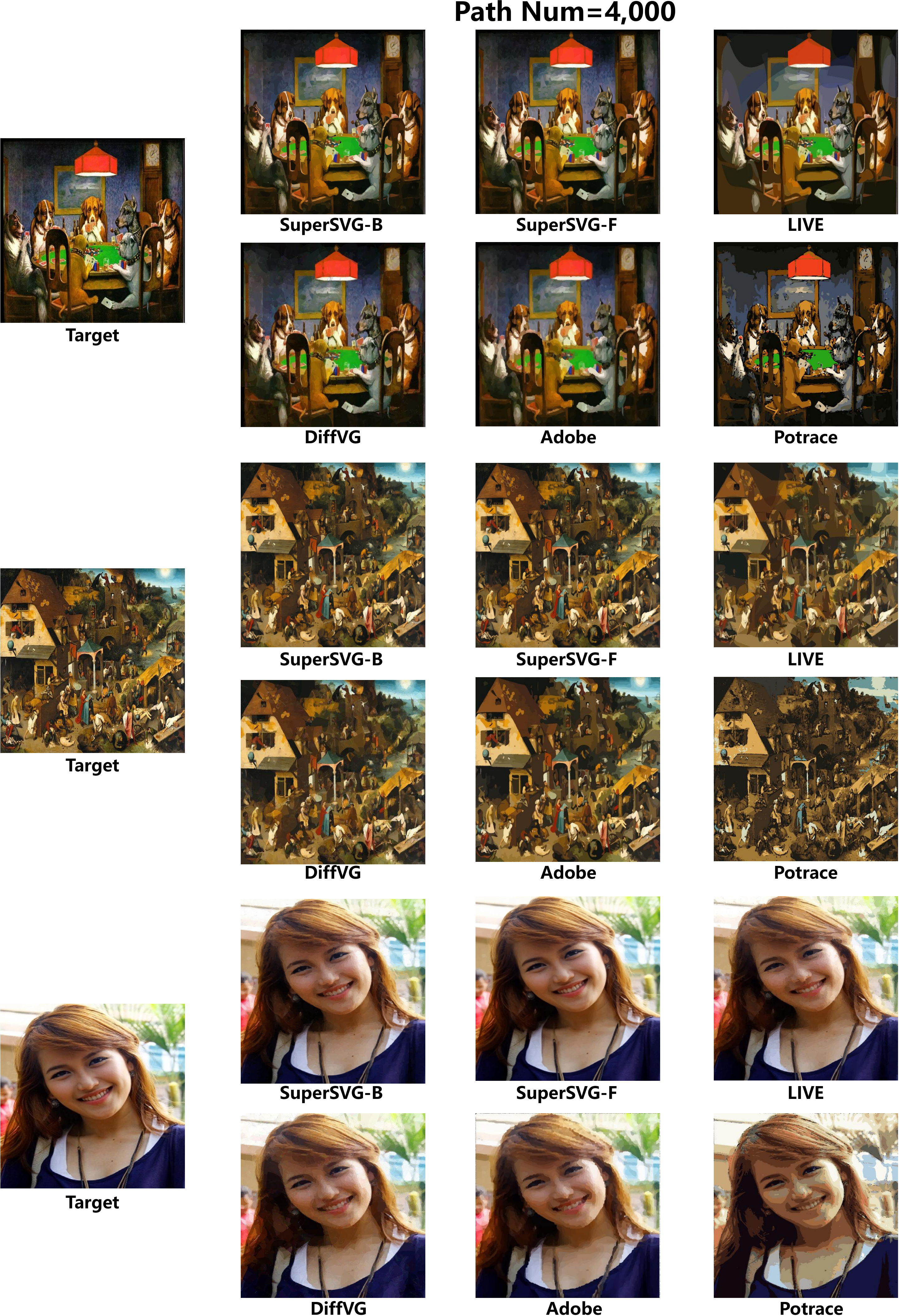}
\caption{More comparison with the state-of-the-art methods \yr{under} 4,000 \yr{SVG} paths.}
\label{fig:more comparison-4000}
\end{figure*}

In this section, we show more comparison results with the state-of-the-art \yr{vectorization} methods, LIVE~\cite{live}, DiffVG~\cite{diffvg}, Adobe~\cite{AdobeIllustrator} and Potrace~\cite{potrace} under 500, 2,000 and 4,000 SVG paths. The results are shown in Figures~\ref{fig:more comparison-500}--\ref{fig:more comparison-4000}. 
It can be seen that \yr{under the same number of SVG paths}, our SuperSVG can reconstruct more details than the other methods in both the foreground and the background \yr{regions}.

\section{More Results of Our Method}
\label{sec:more of our results}
\begin{figure*}[t]
\centering
\includegraphics[width=0.98\textwidth]{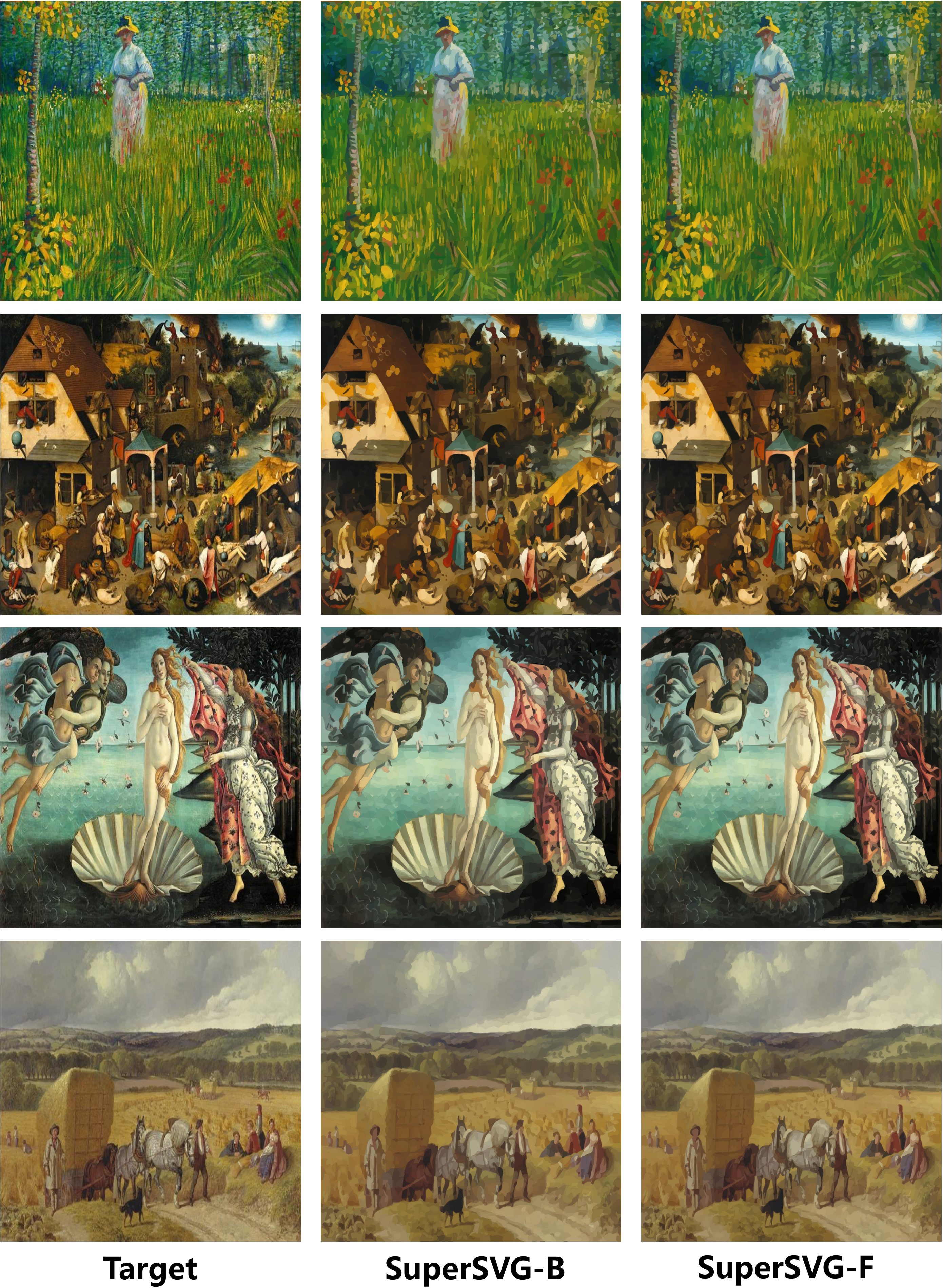}
\caption{More of our results under 4,000 paths. }
\label{fig:more of our result1}
\end{figure*}

\begin{figure*}[t]
\centering
\includegraphics[width=0.98\textwidth]{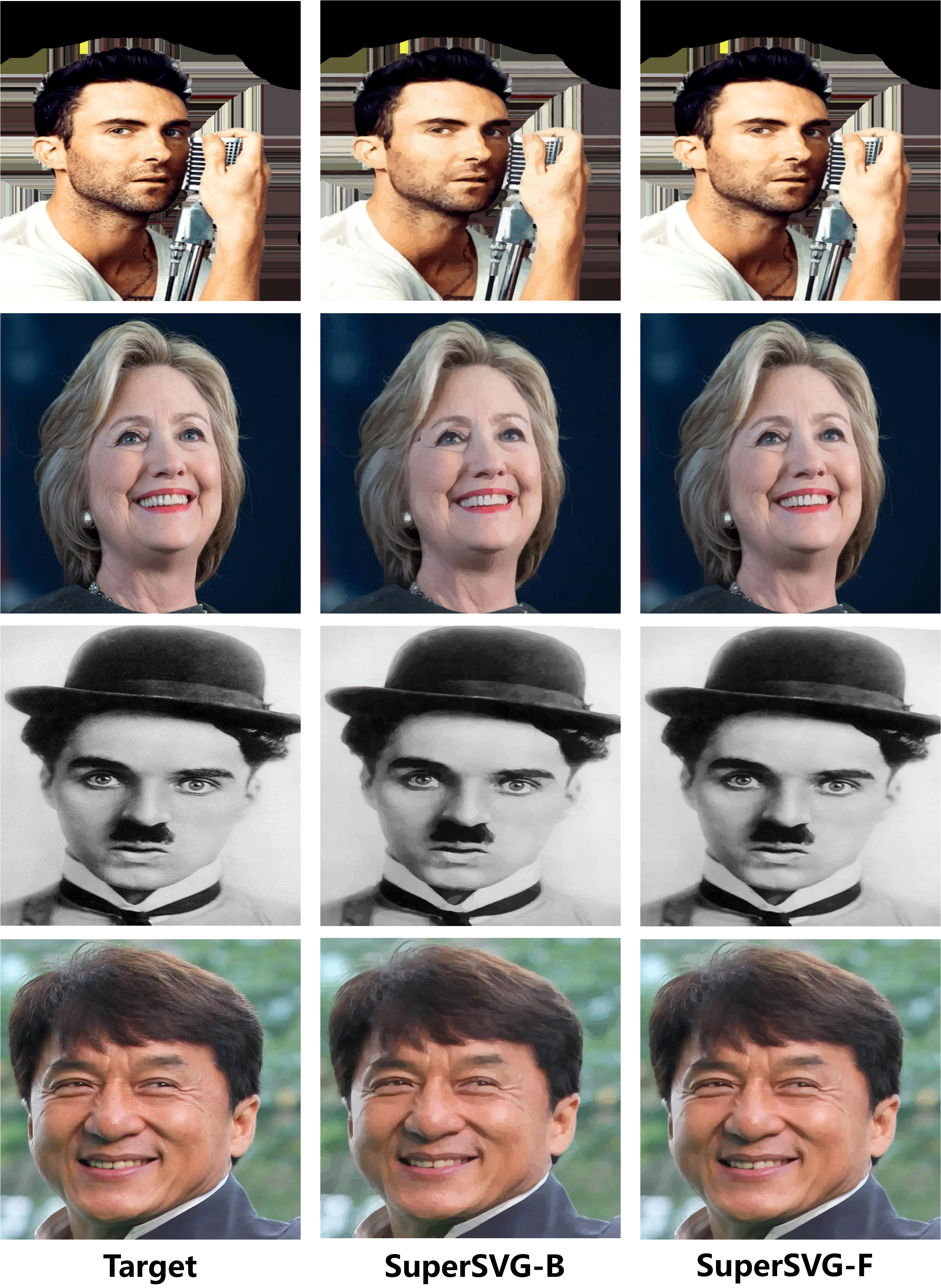}
\caption{More of our results under 4,000 paths. }
\label{fig:more of our result2}
\end{figure*}

\begin{figure*}[t]
\centering
\includegraphics[width=0.98\textwidth]{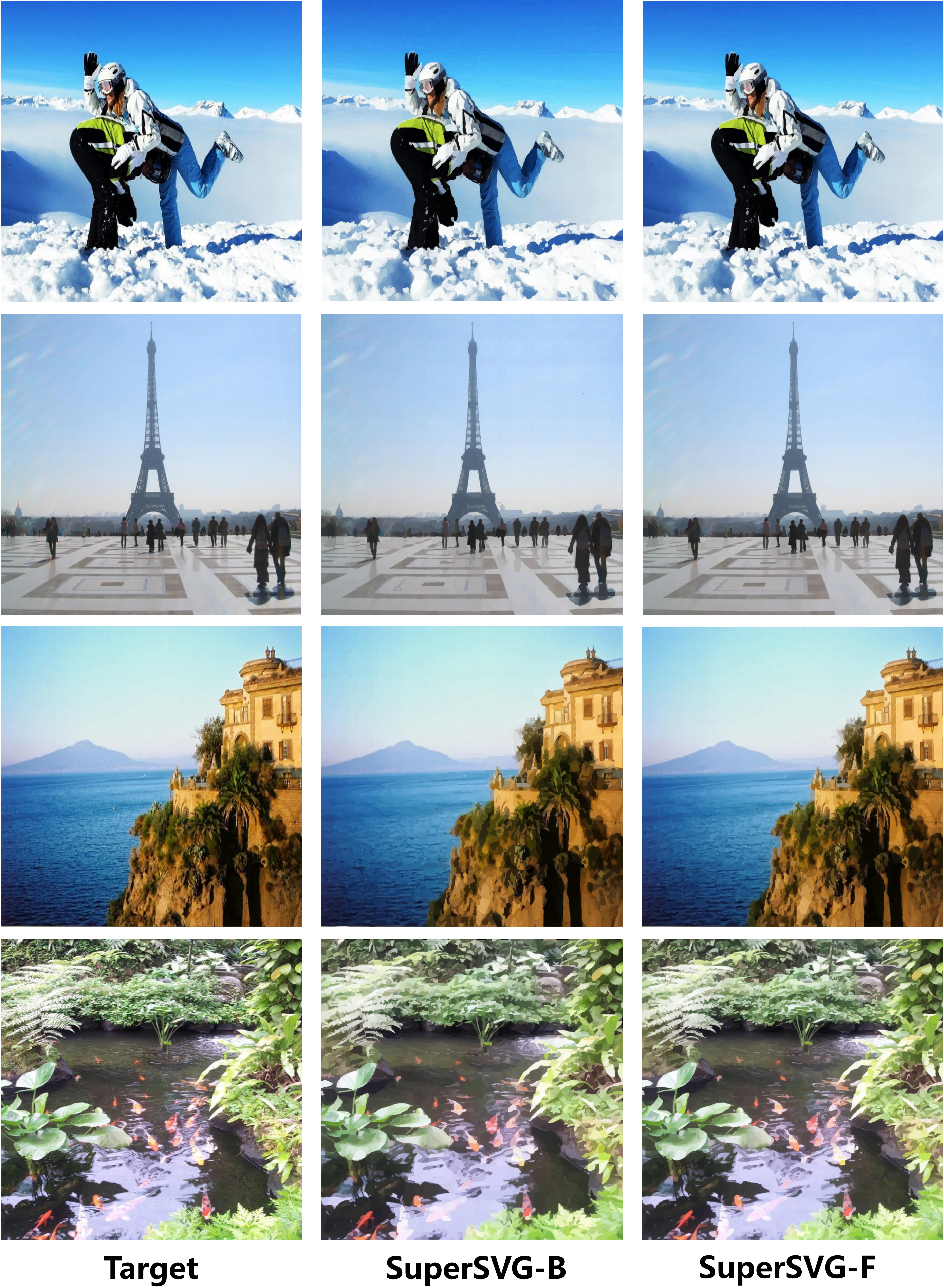}
\caption{More of our results under 4,000 paths. }
\label{fig:more of our result3}
\end{figure*}

\begin{figure*}[t]
\centering
\includegraphics[width=0.98\textwidth]{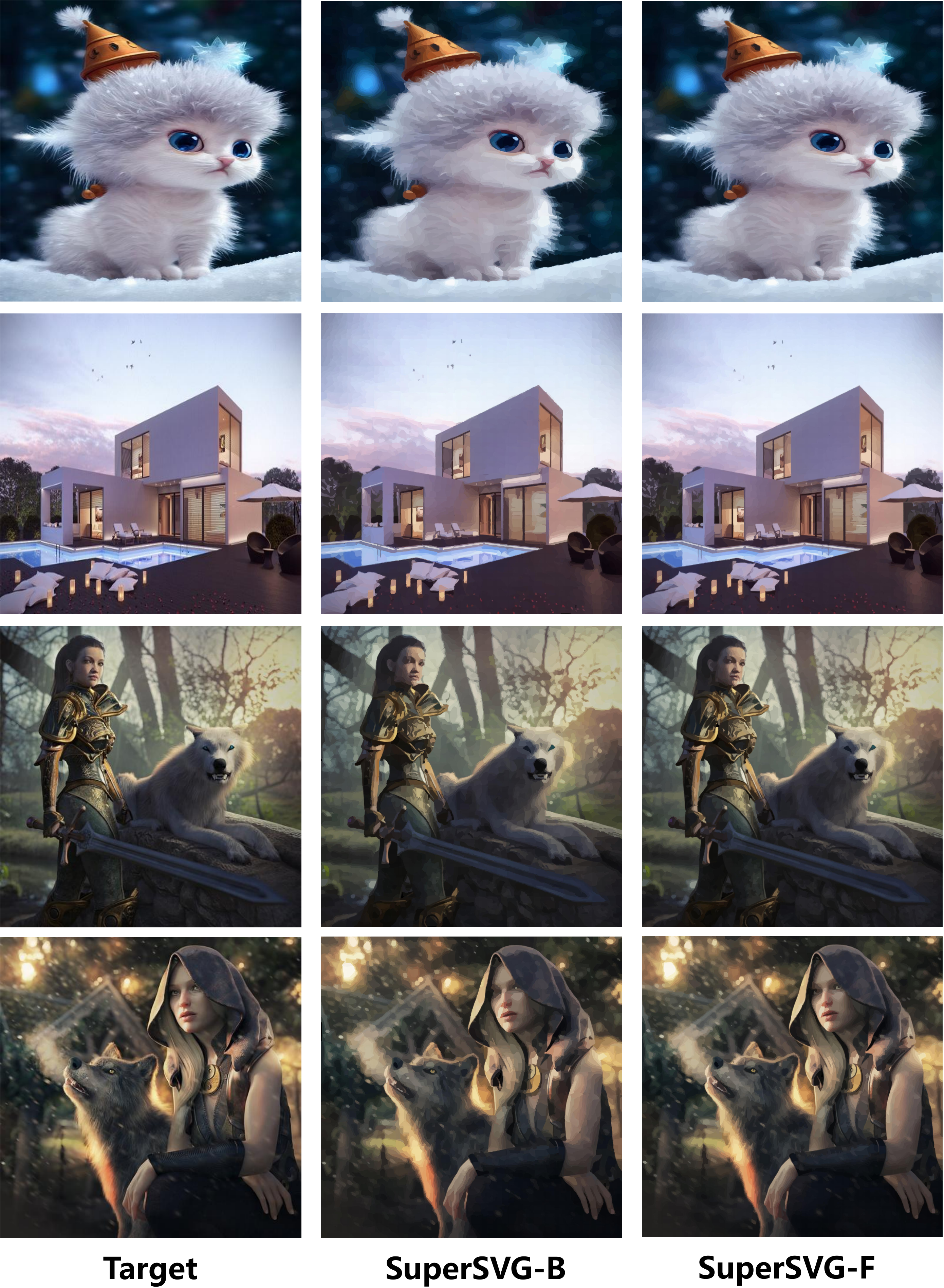}
\caption{More of our results under 4,000 paths. }
\label{fig:more of our result4}
\end{figure*}

To show the effectiveness of our model, we show more experimental results on high-resolution in-the-wild data collected from the Internet~\cite{free-images,pixabay}.
We vectorize all the \yr{test} images into 4,000 SVG paths using \yr{our} SuperSVG-B and SuperSVG-F, \yr{and the results} are shown in Figures~\ref{fig:more of our result1}--\ref{fig:more of our result4}. 
It can be seen that our SuperSVG achieves a good vectorization quality with rich details. 

\newpage

\end{document}